\documentclass{article}
\usepackage[T1]{fontenc}
\usepackage{microtype}
\usepackage{graphicx}
\usepackage{amsmath}
\usepackage{booktabs}
\usepackage{array}
\usepackage{arydshln}
\newcolumntype{M}[1]{>{\centering\arraybackslash}m{#1}}
\newcolumntype{R}[1]{>{\raggedleft\arraybackslash}m{#1}}
\newcolumntype{L}[1]{>{\raggedright\arraybackslash}m{#1}}
\usepackage{placeins}
\usepackage{float}
\usepackage{flafter}
\usepackage{needspace}
\usepackage{xcolor}
\usepackage[hidelinks]{hyperref}
\usepackage[accepted]{mlsys2025}
\hypersetup{pdfsubject={}}
\usepackage{xurl}
\mlsystitlerunning{\ours: Understanding When and Why VLA Inference Latency Matters}

\makeatletter
\renewenvironment{figure*}[1][tp]{\@dblfloat{figure}[#1]\centering\setlength{\linewidth}{5.5in}}{\end@dblfloat}
\AddToHook{env/table/begin}{\small\setlength{\tabcolsep}{3pt}\def\tablecaptiongap{\vspace{4pt}}}
\let\tablecaptiongap\relax
\usepackage{etoolbox}
\apptocmd{\@makecaption}{\tablecaptiongap}{}{\PackageError{main}{Cannot add the table caption gap}{}}
\g@addto@macro\appendix{%
  \renewenvironment{table}[1][tbp]{\@dblfloat{table}[tp]}{\end@dblfloat}}
\makeatother
\AtBeginEnvironment{thebibliography}{\raggedright}
\renewcommand{\FloatBarrier}{}
\makeatletter
\newwrite\layout@out
\immediate\openout\layout@out=\jobname.pos
\newcount\layout@seq
\newif\iflayout@wide
\newif\iflayout@infloat
\newcommand{\layout@mark}[2][]{%
  \global\advance\layout@seq\@ne
  \pdfsavepos
  \edef\layout@tmp{\write\layout@out{\the\layout@seq\space #2\space
    \noexpand\the\c@page\space\noexpand\the\noexpand\pdflastxpos\space
    \noexpand\the\noexpand\pdflastypos\space #1}}%
  \layout@tmp}
\AddToHook{env/figure/begin}{\global\layout@widefalse}
\AddToHook{env/table/begin}{\global\layout@widefalse}
\AddToHook{env/figure*/begin}{\global\layout@widetrue}
\g@addto@macro\appendix{\AddToHook{env/table/begin}{\global\layout@widetrue}}
\AddToHook{cmd/caption/before}{\layout@mark{\iflayout@wide wide\else float\fi}}
\AddToHook{env/figure/begin}{\global\layout@infloattrue}
\AddToHook{env/figure*/begin}{\global\layout@infloattrue}
\AddToHook{env/table/begin}{\global\layout@infloattrue}
\AddToHook{env/figure/end}{\global\layout@infloatfalse}
\AddToHook{env/figure*/end}{\global\layout@infloatfalse}
\AddToHook{env/table/end}{\global\layout@infloatfalse}
\AddToHook{begindocument/end}{%
  \immediate\write\layout@out{0 textbottom 0 0 \number\dimexpr\paperheight
    -1in-\voffset-\topmargin-\headheight-\headsep-\textheight\relax}%
  \NewCommandCopy\layout@ref\ref
  \DeclareRobustCommand\ref{\@ifstar{\layout@ref*}{\layout@refmark}}}
\newcommand{\layout@refmark}[1]{%
  \iflayout@infloat\else\layout@mark[\detokenize{#1}]{ref}\fi\layout@ref{#1}}
\makeatother

\makeatletter
\newcommand{\layout@ifwaiting}[2]{%
  \ifx\@dbldeferlist\@empty\ifx\@deferlist\@empty#2\else#1\fi\else#1\fi}
\let\flushfloats\relax
\g@addto@macro\appendix{\def\flushfloats{\layout@ifwaiting{%
  \par\null\newpage\if@firstcolumn\else\null\newpage\fi
  \layout@ifwaiting{\clearpage}{}}{}}}
\g@addto@macro\appendix{%
  }
\AddToHook{cmd/section/before}{%
  \flushfloats\Needspace{4\baselineskip}\layout@mark{section}}
\AddToHook{cmd/subsection/before}{%
  \Needspace{4\baselineskip}\layout@mark{subsection}}
\newcommand{\questionspace}{\if@nobreak\else\Needspace{5\baselineskip}\fi}
\makeatother
\newcommand{\finding}[2]{\par\nopagebreak\noindent\textbf{Finding~#1.}\hspace{2em}\emph{#2}\par}
\makeatletter
\renewcommand{\subparagraph}{\@startsection{subparagraph}{5}{1em}{1.5ex plus 0.5ex minus .2ex}{-2em}{\normalsize\bf}}
\makeatother
\newcommand{\ours}{\textsc{VLALagBench}}

\begin{document}
\sloppy

\twocolumn[
\mlsystitle{\ours: Understanding When and Why\\ VLA Inference Latency Matters}
\begin{mlsysauthorlist}
\mlsysauthor{Qi Wen}{nus}
\mlsysauthor{Borui Xu}{nus}
\mlsysauthor{Hussein Amro}{nus}
\mlsysauthor{Hongshi Tan}{nus}
\mlsysauthor{Hanmin Lin}{ntu}
\mlsysauthor{Chaoyi Ruan}{nus}
\mlsysauthor{Shanghao Liu}{nus}
\mlsysauthor{Gang Yang}{nus}
\mlsysauthor{Yuejiang Liu}{nus,stanford}
\mlsysauthor{Jiafei Duan}{nus}
\mlsysauthor{Bingsheng He}{nus}
\mlsysauthor{Jason Clemons}{nvidia}
\mlsysauthor{Christos Kozyrakis}{stanford,nvidia}
\mlsysauthor{Wenqi Jiang}{nus}
\par\vspace{4pt}
{\small \textsuperscript{1}National University of Singapore\quad \textsuperscript{2}Nanyang Technological University\quad \textsuperscript{3}Stanford University\quad \textsuperscript{4}NVIDIA}
\end{mlsysauthorlist}
\mlsysaffiliation{nus}{National University of Singapore}
\mlsysaffiliation{ntu}{Nanyang Technological University}
\mlsysaffiliation{stanford}{Stanford University}
\mlsysaffiliation{nvidia}{NVIDIA}
\mlsyskeywords{Vision-language-action models, latency, dynamic manipulation, benchmark}
\vskip 0.3in
\begin{abstract}
Vision-language-action (VLA) models offer a promising path toward general-purpose robot manipulation.
While inference latency can affect task success in a constantly evolving physical world, the relationship between latency and task success rate remains poorly understood.
In this paper, we introduce \ours{}, a comprehensive benchmark for studying VLA inference latency sensitivity across task types, object speeds, execution modes (synchronous and asynchronous), and VLA policies.
Using \ours{}, we find that even a modest inference latency of 100\,ms can nearly halve task success on moving targets, with sensitivity varying substantially across tasks and policies.
These differences reflect task-specific requirements for precision and continuous motion.
By overlapping inference with action execution, asynchronous execution can recover some of the lost success at the cost of more frequent inference requests.
Building on these findings, we develop an empirical model that predicts task success from limited calibration data, achieving root-mean-square prediction errors of only 5.7--7.8\% across various VLA policies.
Given an inference system's latency and per-robot throughput, these predictions guide the selection of system-specific VLA algorithm configurations to maximize task success.
We believe \ours{} and these findings provide a solid foundation for future robotics algorithm--system co-design.
\end{abstract}

\vskip 0.15in
]
\makeatletter\chead{\small\bf\@mlsystitlerunning}\makeatother

\setlength{\footskip}{20pt}
\fancyfoot[C]{\thepage}
\thispagestyle{plain} %

\sloppy
\section{Introduction}
\label{sec:introduction}

Physical AI has the potential to enable robots to perform diverse tasks autonomously in the real world.
Recent advances in vision-language-action (VLA) models have brought this goal closer by predicting robot actions via visual observations and natural-language instructions.
These models have demonstrated strong capabilities across a wide range of manipulation tasks, highlighting their potential for general-purpose robotics~\citep{intelligence2025pi05,nvidia2025gr00tn1,fang2026molmoact2}.

Although VLA models need to respond in real time to keep up with changes in the physical world, the inference latency required for successful task execution remains poorly understood.
From a systems perspective, VLA-Perf~\cite{jiang2026vlaperf} uses 10\,Hz and 100\,Hz as reference points for acceptable and high-performance inference, respectively, without evaluating how these rates translate into task success.
Similarly, Realtime-VLA~\cite{ma2025realtimevla,yang2026realtimevlav2,niu2026realtimevlaflash} improves inference and execution speed, but demonstrates the benefits on a limited set of tasks, such as catching a falling pen.
From a benchmarking perspective, popular robotic manipulation benchmarks such as LIBERO~\cite{liu2023libero} and RoboCasa~\cite{nasiriany2024robocasa} primarily evaluate model generalization across tasks and scenes, but their static environments provide limited insight into how inference latency affects task success in evolving physical environments.
While recent benchmarks such as DynamicVLA's DOM, MoveBench, and ReflexBench~\citep{xie2026dynamicvla,zhang2026ppc,chen2026reflex} support manipulation tasks with moving objects, these benchmarks do not systematically evaluate how inference latency affects task success: their evaluations either omit latency, use latencies tied to specific hardware, or cover only a limited set of policies and tasks.

\begin{figure*}[t]
  \centering
  \includegraphics[width=\textwidth]{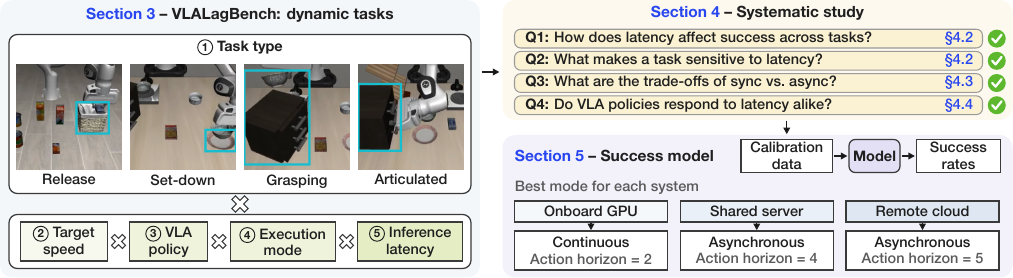}
  \vspace{-2em}
  \caption{\ours{} links VLA inference latency to task success through observation staleness and idle time, enabling a compact model to predict success at unseen latencies and choose the execution mode for a given deployment.
  The benchmark varies five factors---task type, target speed, inference latency, execution mode, and policy---to answer Q1--Q4. }
  \label{fig:overview}
  \vspace{-1em}
\end{figure*}

In this paper, \textbf{we present \ours{}, a benchmark for systematically studying how VLA inference latency affects task success}, which provides the foundation for future robotic system design.
As shown in Figure~\ref{fig:overview}, \ours{} examines five factors related to latency sensitivity.
(1) \emph{Task type}: grasping a moving object, releasing an object into a moving receptacle, placing an object on a moving surface, and manipulating a moving articulated object.
(2) \emph{Object speed}: different target speeds within each task.
(3) \emph{VLA policy}: popular models including $\pi_{0.5}$, GR00T N1.7, and MolmoAct2~\citep{intelligence2025pi05,nvidia2026gr00tlibero,fang2026molmoact2}.
(4) \emph{Execution mode}: synchronous and various asynchronous execution methods.
(5) \emph{Inference latency}: controlled latency levels.
Beyond our primary benchmark built on the MuJoCo simulator~\cite{todorov2012mujoco}, we also conduct experiments in Isaac Sim  (another widely-used simulator) and on real robots to verify our findings.

Our evaluation with \ours{} yields four \textbf{key findings}, each addressing a research question:\par \smallskip \begingroup
\setlength{\parindent}{0pt}
\setlength{\parskip}{4pt}
\textbf{Q1.
How does inference latency affect task success across different manipulation tasks?}\newline
Latency sensitivity varies substantially across tasks (\S\ref{sec:eval-tasks}).
For objects moving at $0.1$\,m/s, a moderate inference latency of 100\,ms nearly halves success in one task (from 71.5\% to 40.5\%), while another retains more (86\%) of its zero-latency success rate (from 86.5\% to 74.0\%).\par

\textbf{%
Q2.
What factors determine a task's sensitivity to inference latency?}\newline
Latency reduces task success through (1) stale observations (the target has moved by the time inference completes) and (2) robot pauses during synchronous inference.
Their impact varies by task: stale observations affect tasks that particularly require precise manipulation, while pauses disrupt tasks requiring continuous motion (\S\ref{sec:eval-tasks}).
These factors motivate our next study of asynchronous execution.\par

\textbf{%
Q3.
What are the trade-offs between synchronous and asynchronous execution as inference latency increases?}\newline
Asynchronous execution overlaps inference with robot actions, keeping the robot moving and benefiting tasks that require continuous motion.
When inference latency is sufficiently short, continuous asynchronous execution also reduces observation staleness through more frequent action updates, improving precision.
These benefits come at the cost of more frequent inference requests (\S\ref{sec:eval-async}).
\par
\textbf{Q4.
Do different VLA models exhibit the same latency sensitivity?}\newline
VLA policies differ in their sensitivity to stale observations and robot pauses, thus benefiting to different extents from asynchronous execution.
Additionally, the best-performing policy at a given inference latency depends on both the task and target speed (\S\ref{sec:eval-policies}).\par \endgroup \smallskip

Building on these findings, \textbf{we further develop an empirical model that guides the selection of algorithm configurations for a given inference system} (\S\ref{sec:eval-deployment}).
Different combinations of GPUs, networks, and software runtimes result in different inference latencies and per-robot throughput.
Consequently, the VLA configuration, including the execution mode and action execution horizon, that maximizes task success may differ across systems.
Because exhaustively evaluating this configuration space is prohibitively expensive, we develop a model that estimates task success across latencies, execution modes, and execution horizons, using only a small set of measurements.
Given a system's latency and throughput constraints, we thus use these predictions to select the feasible algorithm configuration with the highest predicted task success.

\Needspace{6\baselineskip} This paper makes three main \textbf{contributions}, which together provide a solid foundation for future robotics algorithm--system co-designs:
\begin{itemize}
  \setlength{\itemsep}{2pt}
  \setlength{\parsep}{0pt}
  \item {We introduce \ours{} to comprehensively benchmark how VLA latency affects task success.}
  \item {We present findings on latency sensitivity across tasks, target speeds, execution modes, and VLA policies.}
  \item {We develop an empirical model that predicts task success rate and use it to guide algorithm configuration choice given a user-provided system specification.}
\end{itemize}

\begin{figure}[t]
  \centering
  \includegraphics[width=\linewidth]{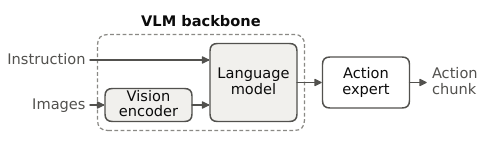}
  \vspace{-2.5em}
  \caption{A VLA model combines a VLM backbone with an action expert that predicts a sequence of robot actions.}
  \label{fig:vla-architecture}
\end{figure}

\section{Background and Motivation}
\label{sec:background}

In this section, we introduce VLA models and review existing benchmarks for evaluating their manipulation performance.
Existing benchmarks offer limited insight into how inference latency affects task success, motivating our systematic investigation on this important topic.

\subsection{Vision-Language-Action Models}

Vision-language-action (VLA) models represent an important step toward physical AI by using visual observations and natural-language instructions to generate robot actions for continuous control~\citep{black2024pi0,intelligence2025pi05,nvidia2025gr00tn1}.
As shown in Figure~\ref{fig:vla-architecture}, a VLA model combines a vision-language model (VLM) backbone that encodes camera images and language instructions with an action expert that predicts the robot's next sequence of actions.
Models such as $\pi_0$, $\pi_{0.5}$, GR00T N1, and MolmoAct2 use diffusion transformers (DiT) as action experts, transforming sampled noise into continuous action sequences through multiple refinement iterations~\citep{black2024pi0,intelligence2025pi05,nvidia2025gr00tn1,fang2026molmoact2}.
OpenVLA and the original MolmoAct instead generate discrete action tokens autoregressively, which are then decoded into robot actions~\citep{kim2024openvla,lee2025molmoact}.

\begin{figure}[t]
  \centering
  \includegraphics[width=\linewidth]{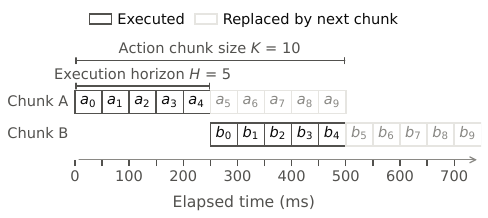}
  \vspace{-2.5em}
  \caption{Each VLA inference generates a chunk of actions.}
  \label{fig:action-chunking}
\end{figure}

Modern VLA policies commonly predict an \emph{action chunk}, a sequence of future actions generated in one inference call~\citep{zhao2023act,black2024pi0,nvidia2025gr00tn1,fang2026molmoact2}.
We denote the number of predicted actions by $K$, the \emph{action chunk size}, and the number executed before switching to an updated chunk by $H$, the \emph{execution horizon}~\citep{black2025rtc}.
Predicting multiple actions together helps maintain consistent motion, while executing only a shorter prefix ($H<K$) lets the robot incorporate new observations and adjust its actions sooner~\citep{chi2023diffusionpolicy,black2025rtc}.
Figure~\ref{fig:action-chunking} shows an example of two chunks generated by separate inference calls (assuming zero inference latency): the robot executes the first $H=5$ actions of chunk A ($K=10$), then switches to chunk B, replacing A's remaining actions.

\subsection{Robot Manipulation Benchmarks}

Widely used robotic manipulation benchmarks evaluate how well policies perform diverse tasks and generalize across objects, goals, and scenes.
For example, LIBERO provides task suites that vary spatial relationships, object identities, and task goals, together with a long-horizon suite that requires executing sequences of manipulation skills~\citep{liu2023libero}.
RoboCasa expands evaluation to diverse kitchen environments, covering individual skills such as pick-and-place and opening drawers, as well as composite activities that combine multiple skills~\citep{nasiriany2024robocasa}.
RoboLab evaluates generalist policies on household manipulation tasks organized around visual recognition, spatial relationships, and procedural skills, and measures their robustness to changes in scene appearance and camera configuration~\citep{yang2026robolab}.

These benchmarks provide valuable measures of task competence and generalization, but limited insight into the consequences of inference latency.
Due to the static physical environments in these benchmarks, a longer inference latency increases task completion time without directly affecting the task outcome~\citep{chen2026reflex}.
Consequently, these benchmarks do not reflect whether a policy can react quickly enough to succeed when the physical world continues to evolve during inference.

\subsection{Dynamic Manipulation Benchmarks}

Recent dynamic manipulation benchmarks introduce moving objects to test policy reactivity, but they do not systematically isolate how inference latency degrades task success.
For example, MoveBench tests grasping under constant, accelerating, and irregular object velocities~\citep{zhang2026ppc}, DynamicVLA's Dynamic Object Manipulation (DOM) benchmark evaluates grasping and placing with moving targets~\citep{xie2026dynamicvla}, and ReflexBench introduces time-critical tasks such as conveyor picking and ball catching~\citep{chen2026reflex}.
However, MoveBench does not account for inference latency during simulation~\citep{zhang2026ppc}.
DOM runs the simulator during inference, but its latencies are tied to the policy and GPU used for evaluation, without a controlled sweep showing how task success changes with inference latency~\citep{xie2026dynamicvla}.
ReflexBench explicitly varies inference latency under synchronous and asynchronous execution, but studies its effect using a single policy (SmolVLA) on only two tasks~\citep{chen2026reflex}.

\begin{figure*}[t]
  \centering
  \includegraphics[width=\textwidth]{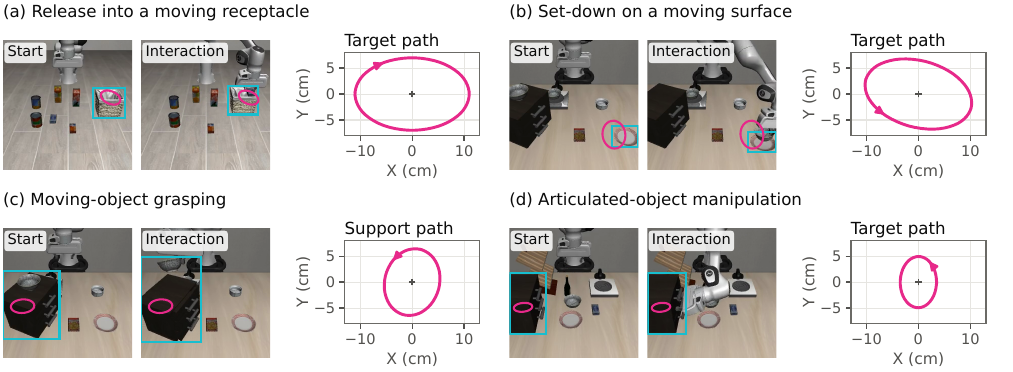}
  \vspace{-2.5em}
  \caption{ The four primary tasks in \ours{} cover releasing, setting down, grasping, and articulated-object manipulation.
  Each target, highlighted by a cyan box, moves continuously along the closed path shown in pink without colliding with background objects. }
  \vspace{-1em}
  \label{fig:tasks}
\end{figure*}

\textbf{Research gap.} Because neither existing benchmarks nor latency-aware model papers (Appendix~\ref{app:related}) jointly vary task types, object speeds, execution modes, and generalist VLA policies under controlled inference latency, the systematic relationship between VLA latency and robot task success rate remains uncharacterized.

\section{Benchmark and Execution Modes}
\label{sec:benchmark}

We present \ours{}, a benchmark for systematically studying how VLA inference latency affects task success.
As summarized in Figure~\ref{fig:overview}, we vary five factors: (1) manipulation task type, (2) target object speed, (3) VLA policy, (4) execution mode, and (5) inference latency.
We hope the insights from the systematic evaluation of \ours{} (\S\ref{sec:evaluation}) will guide future robotic system and architecture designs.

\subsection{Design Principles}
\label{sec:design}

To ensure that performance drops reflect latency sensitivity rather than baseline policy failures or simulator artifacts, \ours{} is built around four design principles.
First, evaluated policies must achieve high success rate in static scenes at zero latency so that accuracy degradation with added latency is attributable to latency.
Second, the tasks should cover diverse interactions with moving targets, such as grasping a moving object or placing an object at a moving destination.
Third, object moving trajectories must be precisely controllable without altering the simulator's underlying contact physics.
Fourth, multiple widely adopted VLA models must provide public checkpoints on the exact same tasks to enable cross-policy comparisons.

These principles rule out most of the widely used simulated manipulation benchmarks we screened.
The first principle alone excludes several benchmarks because public VLA checkpoints have limited success even in static scenes.
For example, $\pi_{0.5}$ averages only 16.9\% success on RoboCasa365~\citep{nasiriany2026robocasa365} and 28.0\% on RoboLab~\citep{yang2026robolab}.
Similarly, $\pi_0$ averages 47.9\% on Meta-World~\citep{yu2019metaworld,shukor2025smolvla} and 46.4\% on the clean scenes of RoboTwin~2.0's bimanual tasks~\citep{chen2025robotwin2}, while GR00T N1.7 averages 62.3\% on SimplerEnv's WidowX scenes~\citep{li2024simpler,nvidia2026gr00tsimpler}.
Kinetix~\citep{matthews2024kinetix} fails the second and fourth principle: its 2D physics control tasks are evaluated using small policies trained per environment rather than VLA models~\citep{black2025rtc}.

Thus, we build our primary benchmark on LIBERO in MuJoCo, which satisfies our design principles and allows controlled target motion while preserving realistic contact physics~\citep{todorov2012mujoco}.
We use DOM in Isaac Sim for secondary cross-simulator verification~\citep{xie2026dynamicvla,nvidiaisaacsim}\ (Appendix~\ref{app:dynamicvla}).
We also verify our findings through real-robot experiments (Appendix~\ref{app:real-robot}).

\subsection{Tasks}
\label{sec:tasks}

We select four representative tasks that cover distinct ways a robot interacts with a moving target (Figure~\ref{fig:tasks}).

\vspace{-.3em}
\noindent\textbf{(a) Release.} The robot picks up a stationary cream cheese package and drops it into a moving basket.

\vspace{-.3em}
\noindent\textbf{(b) Set-down.} The robot picks up a stationary bowl and places it on a moving plate.

\vspace{-.3em}
\noindent\textbf{(c) Grasping.} The robot picks up a bowl carried by a moving cabinet and places it on a stationary plate.

\vspace{-.3em}
\noindent\textbf{(d) Articulated manipulation.} The robot grasps the handle of the moving cabinet's middle drawer and pulls it open.

We select these four tasks because they achieve near-perfect policy success in static scenes, show distinct robot--object interactions, and are sensitive to target motion.
Appendix~\ref{app:task-selection} explains why other LIBERO tasks were not selected.

These tasks differ in how the robot interacts with the moving target.
\emph{Release} and \emph{set-down} deliver an object to a moving destination, whereas \emph{grasping} and \emph{articulated-object manipulation} require the robot to acquire a moving object or handle directly.
For the moving destination tasks, release allows coarse alignment over a wide basket opening, whereas set-down requires precise placement near the center of a narrow plate.
For the moving object tasks, grasping requires intercepting the bowl and lifting it from its moving support, whereas articulated-object manipulation requires sustained contact and motion to pull the drawer open.

Each task is engineered so that driving the target along a trajectory preserves realistic contact physics upon robot interaction.
For \emph{release} and \emph{set-down}, we kinematically update the horizontal pose of the basket and plate; because the robot drops the cream cheese from above in \emph{release}, it avoids rigid collisions with the unyielding basket walls.
For \emph{grasping}, kinematically locking the bowl would prevent the robot from lifting it, so we instead drive the supporting cabinet beneath the bowl and let surface friction carry the bowl along the path.
For \emph{articulated-object manipulation}, we move the entire cabinet body while leaving the drawer's damped prismatic joint fully physical.

To validate our findings in a second simulator, we evaluate release and grasping in Isaac Sim using the Dynamic Object Manipulation (DOM) benchmark and the accompanying DynamicVLA policy trained on moving objects~\mbox{\citep{xie2026dynamicvla}}.
In this secondary suite (Figure~\ref{fig:dynamicvla-tasks}, Appendix~\ref{app:dynamicvla}), the robot either drops a can into a moving box (\emph{release}) or picks a can off a moving invisible rail carrier to place into a static box (\emph{grasping}).
However, DOM's native tasks are largely homogeneous---primarily picking up a moving object and placing it into a static receptacle---and the only model supporting this benchmark with a public checkpoint is DynamicVLA, which fails to adapt to new task setups and exhibits persistent oscillatory movements (acting as if intercepting a fast-moving target even when the object rests completely still).
We therefore use LIBERO tasks as the primary benchmark and report DynamicVLA as a cross-simulator verification in Appendix~\ref{app:dynamicvla}.

\subsection{Motion Parameters}
\label{sec:motion}

Each target moves at a constant speed along a closed, nearly elliptical path that avoids collisions with background objects in LIBERO's original layouts.
We sweep target speeds from $0.001$\,m/s (near-static) to $0.2$\,m/s across all four tasks; Appendix~\ref{app:protocol} provides trajectory and initialization details.

\begin{figure}[t]
  \centering
  \includegraphics[width=\linewidth]{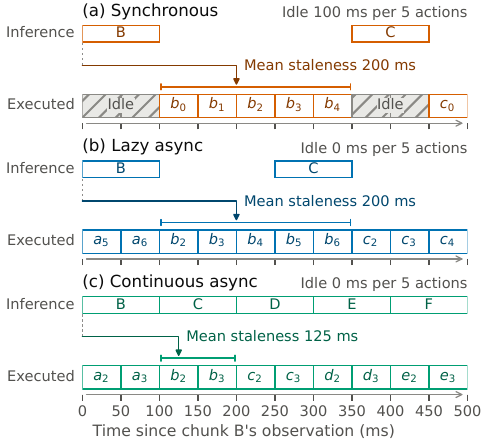}
  \vspace{-2.5em}
  \caption{Synchronous and two types of asynchronous execution, assuming 100\,ms inference latency and 50\,ms per action.}
  \label{fig:schedule-quantities}

\end{figure}

\subsection{VLA Policies}
\label{sec:vla-policies}

\ours{} evaluates three representative VLA policies: $\pi_{0.5}$, GR00T N1.7, and MolmoAct2, using their official public checkpoints~\citep{openpi2025libero,nvidia2026gr00tlibero,fang2026molmoact2}.
These checkpoints achieve near-perfect success in static scenes, providing a reliable baseline for comparing latency sensitivity across policies.
For cross-simulator verification on DOM in Isaac Sim, we evaluate DynamicVLA, a policy explicitly trained on moving objects on the DOM benchmark~\citep{xie2026dynamicvla}.

\subsection{Execution Modes}
\label{sec:execution-modes}

We compare synchronous execution with two representative asynchronous schedules: lazy and continuous execution (Figure~\ref{fig:schedule-quantities}).
An execution mode governs when the robot requests a new action chunk and how it acts while waiting, creating a fundamental trade-off between robot idle time, observation staleness, and inference frequency.
Across the three modes, $K$ denotes the number of predicted actions per chunk, $H$ the number executed per chunk, the execution horizon, and $L$ the inference latency expressed as a number of control steps (\S\ref{sec:latency-injection}).

\paragraph{Mode 1: Synchronous execution.}
Synchronous execution is a common deployment pattern in which inference and robot action execution interleaves~\citep{black2025rtc}.
After executing $H$ actions, the robot requests a new chunk and pauses for $L$ control steps while inference completes, holding its position and last gripper command.
Once the action chunk becomes available, the robot resumes execution from its first predicted action.

\textit{Pros and cons.} Its advantage is its low inference request rate: one request every $H+L$ control steps.
Its disadvantage is that the robot pauses during every inference call, which can disrupt tasks that require continuous motion.

\textit{Idle time.}
We quantify robot pauses using idle time ($I$), the number of control steps spent waiting for inference per $H$ executed actions.
Synchronous execution incurs $I=L$ idle steps per execution horizon (hatched in Figure~\ref{fig:schedule-quantities}).
Once the first action chunk is available, both asynchronous modes introduced below eliminate pauses ($I=0$), provided the available actions cover the next inference interval.

\paragraph{Mode 2: Lazy asynchronous execution.}
Asynchronous execution avoids robot pauses by running inference while the robot executes actions from an earlier chunk.
Lazy execution requests inference least frequently among various asynchronous schedules, starting the next call only when $L$ of the $H$ scheduled actions remain so that the next chunk becomes available as the current execution horizon ends.
When the new chunk arrives, the robot skips its first $L$ actions, which correspond to the elapsed inference interval, and executes the following $H$ actions.

\textit{Pros and cons.} Its advantage is that it keeps the robot moving continuously without pauses.
Its disadvantages are that (1) more frequent inference does not reduce average observation staleness relative to synchronous execution, and (2) robot motion during inference can introduce additional uncertainty for tasks requiring precise manipulation.

\textit{Observation staleness.}
We define observation staleness as the average time elapsed between capturing a visual observation and executing an action generated from it.
As illustrated in Figure~\ref{fig:schedule-quantities}, synchronous execution runs actions $0$ through $H-1$ of each chunk, yielding an average staleness of $(H-1)/2+L$ with an inference latency of $L$ steps.
Because lazy asynchronous execution skips the first $L$ actions, its observation staleness is identically $(H-1)/2+L$ steps.

\paragraph{Mode 3: Continuous asynchronous execution.}
Continuous execution represents the opposite extreme in asynchronous execution, running inference as frequently as possible by starting each new call as soon as the previous one completes, while the robot continues executing previously predicted actions.
At each completion, the robot switches to the new chunk, skips its first $L$ actions, and executes the next $L$ actions while another inference call runs, yielding an execution horizon of $H=L$ and an average observation staleness of $(3L-1)/2$ steps.

\textit{Pros and cons.} Its advantage is more frequent action updates and thus a lower observation staleness than in either of the other modes when latency is shorter than their execution horizon ($L<5$).
The average staleness is even lower than in zero-latency synchronous execution with $H=5$ when $L=1$.
Its disadvantages are a higher inference request rate and more frequent transitions between chunks, which can introduce discontinuities in the robot's actions.

\subsection{Latency Injection Mechanism}
\label{sec:latency-injection}

We simulate inference latency by converting it into a number of simulator control steps.
For example, at a control frequency of 20\,Hz, a latency of 100\,ms corresponds to $L=2$ steps: synchronous execution inserts two stationary actions that hold the robot's position and last gripper command, whereas asynchronous execution continues executing available actions from the previous chunk.
We extend each task's deadline by the number of inference-induced idle steps to preserve its action budget (Appendix~\ref{app:protocol}).

\section{Evaluation and Findings}
\label{sec:evaluation}

In this section, we use \ours{} to (1) examine how inference latency affects success across tasks, (2) identify the factors leading to accuracy loss, (3) assess the trade-offs between synchronous and asynchronous execution, and (4) compare latency sensitivity across VLA policies.
We present the main findings here, with the full experimental setup, complete results, failure stages, and validation in Isaac Sim with DynamicVLA (Appendix~\ref{app:dynamicvla}) and on real robots (Appendix~\ref{app:real-robot}) provided in the appendices.

\begin{figure}[t]
  \centering
  \includegraphics[width=\linewidth]{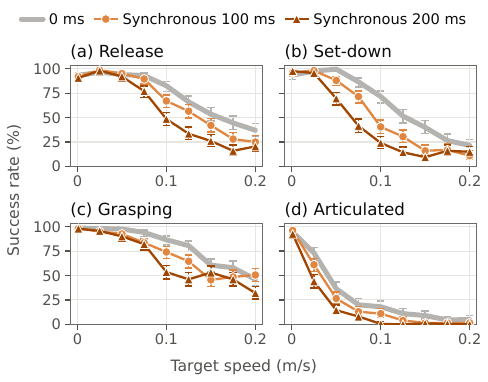}
  \vspace{-2.5em}
  \caption{Task success across target speeds and inference latencies using $\pi_{0.5}$ with synchronous execution.
  Different tasks exhibit distinct latency sensitivities.}
  \label{fig:task-success}
\end{figure}

\subsection{Evaluation setup}
\label{sec:eval-setup}

On LIBERO-based tasks, we test three widely used VLA models, including $\pi_{0.5}$, GR00T N1.7, and MolmoAct2 at a 20\,Hz control rate with an execution horizon of $H=5$ and their default chunk sizes of $K=10$ ($\pi_{0.5}$, MolmoAct2) or $K=16$ (GR00T).
Each reported success rate is based on 200 rollouts, with a budget of 600 executed actions per episode.
This increased budget, up from 220--300 actions in standard static LIBERO, reduces failures caused by insufficient time to pursue moving targets.
We test inference latencies from 0 to 250\,ms in 50\,ms increments because (1) this range covers latencies encountered in practical VLA inference systems~\citep{vlahwbench2026,jiang2026vlaperf}, and (2) at 250\,ms ($L=5$), policies with $K=10$ reach the asynchronous execution limit $K=H+L$.

\FloatBarrier
\begin{figure}[t]
  \centering
  \includegraphics[width=\linewidth]{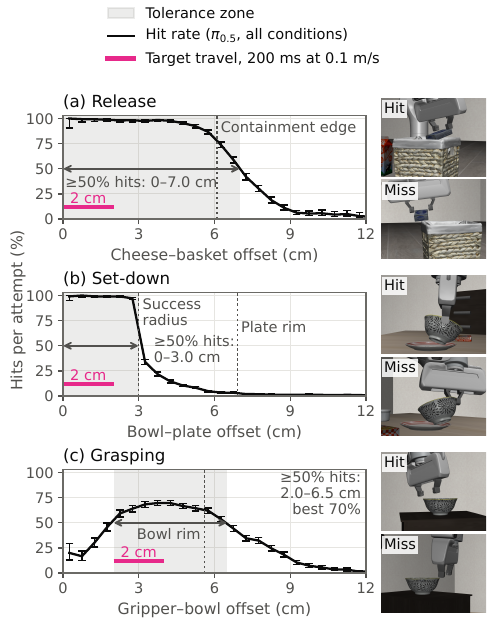}
  \vspace{-1.5em}
  \caption{Hit rate of $\pi_{0.5}$'s attempts against their horizontal offset from the target center when the object leaves the gripper (\emph{release}), first touches the surface (\emph{set-down}), or the fingertips descend below the object's rim (\emph{grasping}); the tolerance zone is the range of offsets where at least half of the attempts hit, and the photos show one hit inside it above one miss outside it.
  \emph{Set-down} requires greater precision than the other tasks.}
  \label{fig:tolerance}
\end{figure}

\subsection{Latency sensitivity across tasks}
\label{sec:eval-tasks}

\questionspace \noindent{{Q1.} \textit{How does inference latency affect task success across different manipulation tasks?}}

\finding{1}{Inference latency can substantially reduce task success on moving targets, with the extent of degradation depending on both the task nature and target speed.}

Empirical success curves confirm that latency penalties emerge primarily when targets move and scale differently across tasks.
As shown in Figure~\ref{fig:task-success}, an inference latency of 200\,ms under synchronous execution changes success on nearly static targets ($0.001$\,m/s) by only a few percentage points across all tasks.
On moving targets, however, even moderate inference latency affects tasks differently: at a target speed of $0.1$\,m/s, 100\,ms of latency nearly halves \emph{set-down} success, from 71.5\% to 40.5\% (57\% of its zero-latency success rate).
In contrast, \emph{grasping} success declines only from 86.5\% to 74.0\%, retaining 86\% of its zero-latency success rate.
We next examine the factors underlying these differences in latency sensitivity across tasks.

\questionspace \noindent{Q2.
\textit{What factors determine a task's sensitivity to latency?}}

\finding{2}{A task's latency sensitivity depends on its requirements for  action precision and continuous motion.
\textit{(2.1)} Tasks requiring high spatial precision suffer from stale observations, whereas \textit{(2.2)} tasks requiring sustained motion suffer from robot pauses.}

{\textit{(2.1) Precision.}}
Stale observations make precise manipulation harder because the target moves between image capture and action execution, so the robot may act on a position the target no longer occupies.
Figure~\ref{fig:tolerance} compares the precision required by different tasks: \emph{set-down} requires placing the bowl within a narrow 3\,cm zone around the plate's center, whereas \emph{release} tolerates drops anywhere across a 12.2\,cm basket opening.
At a target speed of $0.1$\,m/s, 200\,ms of additional observation staleness allows the plate to move 2\,cm, a displacement equal to two-thirds of \emph{set-down}'s tolerance for placement error, making precise placement more difficult.

{\textit{(2.2) Continuous motion.}}
Robot pauses disrupt tasks that require continuous motion, making it harder to catch up with or move alongside the target.
Because VLA policies are trained via imitation learning on static demonstrations~\mbox{\citep{liu2023libero}}, they issue actions at a nearly constant velocity regardless of how fast the target moves (Figure~\ref{fig:hand-speed}, Appendix~\ref{app:protocol}).
Synchronous execution halts the arm after every $H$ actions, reducing $\pi_{0.5}$'s average approach speed by 34\% at 200\,ms latency during \emph{release}.
This reduced speed causes the arm to fall behind the moving basket before releasing, and prevents the gripper from keeping pace with the moving drawer during \emph{articulated-object manipulation}.

As we quantify in \S\ref{sec:eval-model}, \emph{grasping} is more sensitive to observation staleness, \emph{articulated manipulation} and \emph{release} are more sensitive to robot pauses, while \emph{set-down} is sensitive to both.

\begin{figure}[t]
  \centering
  \includegraphics[width=\linewidth]{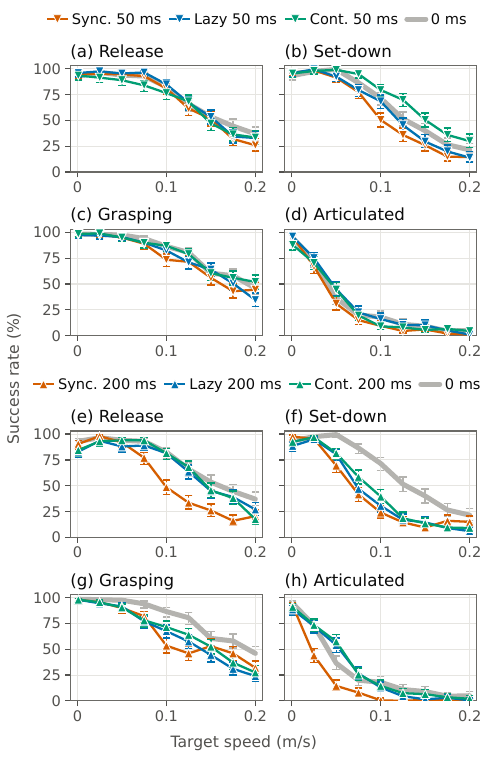}
  \vspace{-2.5em}
  \caption{Task success using $\pi_{0.5}$ under synchronous and asynchronous execution at 50 and 200\,ms inference latency.}
  \label{fig:mode-summary}
\end{figure}

\subsection{Synchronous versus asynchronous inference}
\label{sec:eval-async}

\questionspace \noindent{Q3.
\textit{What are the trade-offs between synchronous and asynchronous execution as inference latency increases?}}

\finding{3}{Asynchronous execution may recover task success by reducing robot pauses and observation staleness, at the cost of more frequent inference (\S\ref{sec:execution-modes}).
\textit{(3.1)} Both lazy and continuous asynchronous execution keep the robot moving during inference, helping tasks that require continuous motion, such as \emph{articulated-object manipulation}.
\textit{(3.2)} At short inference latencies, continuous asynchronous mode updates actions more frequently, helping tasks that require precise manipulation, such as \emph{set-down}.
}

\textit{(3.1) Removing pauses.}
For \emph{release} and \emph{articulated-object manipulation}, the two tasks more sensitive to robot pauses (\S\ref{sec:eval-tasks}), both asynchronous modes help by keeping the robot moving during inference.
Figure~\ref{fig:mode-summary}(a,d,e,h) shows that both modes restore $\pi_{0.5}$'s success close to the zero-latency baseline at both 50 and 200\,ms.
In particular, synchronous and lazy asynchronous execution have the same average observation staleness at the same inference latency (\S\ref{sec:execution-modes}), so the recovery under lazy execution indicates that removing pauses accounts for much of the improvement.

\begin{figure}[t]
  \centering
  \includegraphics[width=\linewidth]{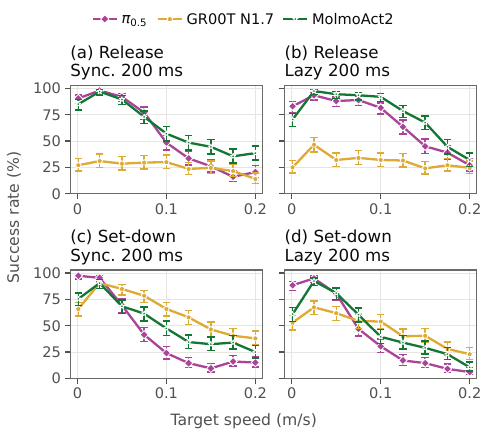}
  \vspace{-2.5em}
  \caption{Task success across VLA policies at 200\,ms latency.
  The best-performing policy depends on the task and target speed.}
  \label{fig:policy-summary}
\end{figure}

\textit{(3.2) Reducing observation staleness.}
For \emph{set-down} and \emph{grasping}, the two tasks more sensitive to stale observations (\S\ref{sec:eval-tasks}), continuous asynchronous execution can improve success by updating actions more frequently when inference latency is sufficiently short.
At 50\,ms ($L=1$), continuous asynchronous execution updates the action chunk at every control step, reducing average observation staleness below that of zero-latency synchronous execution ($H=5$) and increasing \emph{set-down} success at $0.125$\,m/s from 51.5\% to 70.0\% (Figure~\ref{fig:mode-summary}b).
The benefit is more modest for \emph{grasping}: continuous asynchronous outperforms synchronous and lazy execution at several target speeds, but does not consistently exceed the zero-latency baseline (Figure~\ref{fig:mode-summary}c).
As inference latency approaches lazy execution's horizon ($L\to5$), continuous asynchronous updates actions less frequently and its observation staleness approaches that of lazy execution; at 200\,ms, neither asynchronous mode consistently recovers zero-latency success on these two tasks (Figure~\ref{fig:mode-summary}(f,g)).

\begin{table}[t]
  \centering
  \caption{Success rate on nearly static targets ($0.001$\,m/s) across policies.
  Changing initial target positions already substantially reduces success even with zero inference latency.}
  \label{tab:policy-baseline}
  \footnotesize
  \scalebox{1.0}{%
  \begin{tabular}{@{}lrrr@{}}
  \toprule
  Task & $\pi_{0.5}$ (\%) & GR00T N1.7 (\%) & MolmoAct2 (\%) \\
  \midrule
  Release & 93.0 & 52.0 & 83.0 \\
  Set-down & 93.5 & 56.0 & 74.0 \\
  Grasping & 98.5 & 98.0 & 99.5 \\
  Articulated & 94.5 & 69.5 & 81.0 \\
  \bottomrule
  \end{tabular}%
  }
\end{table}

\subsection{Latency sensitivity across VLA policies}
\label{sec:eval-policies}

\questionspace \noindent{Q4.
\textit{Do VLA models exhibit the same latency sensitivity?}}

\finding{4}{{VLA policies differ in their latency sensitivity, so the best-performing policy at a given inference latency depends on both the task and target speed \textit{(4.1)}.
This variation reflects differences in \textit{(4.2)} robustness to changes in initial target position and \textit{(4.3)} sensitivity to observation staleness and robot pauses.}}

\textit{(4.1) Different tasks and speeds favor different policies.}
As shown in Figure~\ref{fig:policy-summary}, $\pi_{0.5}$ achieves the highest success at the slowest target speeds on both \emph{release} and \emph{set-down} under 200\,ms synchronous execution.
At higher speeds, MolmoAct2 performs best on \emph{release}, whereas GR00T N1.7 performs best on \emph{set-down} (Figure~\ref{fig:policy-summary}(a,c)).
The full four-task comparison is shown in Figure~\ref{fig:policy} (Appendix~\ref{app:full-curves}).

\textit{(4.2) Per-policy robustness to initial target position.}
Table~\ref{tab:policy-baseline} shows accuracy of different policies given various initial target positions on nearly static targets ($0.001$\,m/s).
Although all three policies achieve roughly 97\% success on the original static LIBERO layouts, their success rates differ substantially when initial positions are sampled along the motion path.
For example, on \emph{set-down}, $\pi_{0.5}$ achieves 93.5\% success, compared with 74.0\% for MolmoAct2 and 56.0\% for GR00T N1.7.

\textit{(4.3) Per-policy sensitivity to observation staleness and robot pauses.}
VLA policies differ in their sensitivity to observation staleness and robot pauses, leading to different gains from asynchronous execution.
For example, on \emph{release} at a target speed of $0.1$\,m/s and an inference latency of 200\,ms~(Figure~\ref{fig:policy-summary}(a,b)), switching from synchronous to lazy asynchronous execution increases success from 48.5\% to 81.5\% for $\pi_{0.5}$ and from 57.0\% to 92.0\% for MolmoAct2, gains of 33 and 35 percentage points, respectively.
In contrast, GR00T N1.7's success increases only from 30.0\% to 32.0\%, a much smaller gain.
On \emph{set-down}, the same switch can instead reduce GR00T N1.7's success (Figure~\ref{fig:policy-summary}(c,d)).
We quantify these policy-specific sensitivities in \S\ref{sec:eval-model}.

\FloatBarrier
\section{System-Specific Configuration Selection with \ours{}}
\label{sec:eval-deployment}

In this section, we build on the preceding findings to guide VLA configuration selection for a given inference system.
Specifically, we develop an empirical model that uses limited calibration data for each policy--task pair to predict task success across inference latencies, execution modes, and execution horizons (\S\ref{sec:eval-model}).
This model reuses measurements collected on one system to select configurations that maximize predicted task success on a new system (\S\ref{sec:deployment-selection}).

\FloatBarrier
\subsection{An Empirical Model for Task Success Prediction}
\label{sec:eval-model}

\textbf{Success model intuition.}\textit{ Stale observations and robot pauses caused by inference latency can make a task harder, much like increasing the target's speed.
This similarity suggests a way to reuse past success measurements: we first measure how task success varies with target speed under one latency and execution mode.
To estimate success under another  latency and execution modes, we simply look up the success rate on the measured curve at a target speed adjusted for observation staleness and robot idle time.}

\begin{figure}[t]
  \centering
  \includegraphics[width=\linewidth]{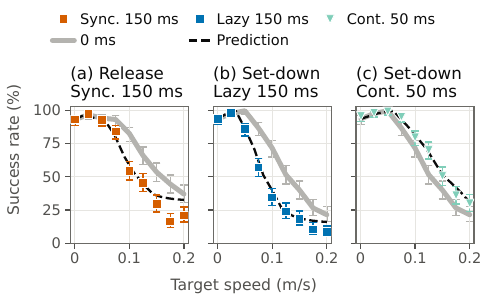}
  \vspace{-2.5em}
  \caption{Model predictions track $\pi_{0.5}$'s measured success at inference latencies excluded from calibration.
  Dashed black curves show predictions; markers show measurements.}
  \label{fig:prediction-summary}
\end{figure}

For a given policy and task, we approximate success under configuration $X$ as $S_X(v)\approx S_0(\lambda_X v)$, where $S_0$ is the success curve under zero-latency synchronous execution with a fixed reference horizon.
Configuration $X$ specifies the inference latency, execution mode, and execution horizon.
The factor $\lambda_X$ adjusts the target speed: to predict success at speed $v$, we read the reference curve at speed $\lambda_X v$.
In practice, we find that this adjustment is well approximated by a simple linear model of observation staleness and robot idle time:
\begin{equation}
  \lambda_X
  =1+a\,\frac{\bar{s}_X-\bar{s}_0}{\bar{s}_0}
  +b\,\frac{I_X}{H_X},
  \label{eq:model}
\end{equation}
where $\bar{s}_X$ and $\bar{s}_0$ are the mean observation staleness under configuration $X$ and the reference configuration, respectively; $H_X$ is the execution horizon, and $I_X$ is the number of idle steps per execution horizon.
The ratio $(\bar{s}_X-\bar{s}_0)/\bar{s}_0$ measures the relative change in observation staleness, while $I_X/H_X$ measures the number of idle control steps per executed action (\S\ref{sec:execution-modes}).
The coefficients $a$ and $b$ are fitted separately for each policy--task pair and quantify its sensitivity to these two factors.

Calibrating the model does not require zero-latency measurements: the reference success curve can come from synchronous or asynchronous execution at a nonzero latency.
For a reference configuration $R$, we use $S_X(v)\approx S_R((\lambda_X/\lambda_R)v)$, where the ratio adjusts target speed relative to $R$.
To estimate the sensitivity to robot pauses (coefficient $b$), the calibration data must include at least one synchronous configuration with nonzero latency.

\begin{table}[t]
  \centering
  \caption{Fitted sensitivities to observation staleness ($a$) and robot idle time ($b$) across tasks and policies.
  $^\dagger$GR00T N1.7 is an outlier with strong sensitivity to initial target positions.}
  \label{tab:sensitivities}
  \footnotesize
  \scalebox{1.0}{%
  \begin{tabular}{@{}lrrrrrr@{}}
  \toprule
   & \multicolumn{2}{c}{$\pi_{0.5}$} & \multicolumn{2}{c}{MolmoAct2} & \multicolumn{2}{c}{GR00T N1.7$^\dagger$} \\
  \cmidrule(lr){2-3}\cmidrule(lr){4-5}\cmidrule(l){6-7}
  Task & $a$ & $b$ & $a$ & $b$ & $a$ & $b$ \\
  \midrule
  \emph{Release} & 0.04 & 0.59 & $-0.09$ & 0.80 & 2.48 & 1.31 \\
  \emph{Set-down} & 0.33 & 0.32 & 0.31 & 0.09 & $-1.00$ & 3.65 \\
  \emph{Grasping} & 0.23 & 0.07 & 0.23 & $-0.06$ & 0.26 & $-0.37$ \\
  \emph{Articulated} & $-0.03$ & 1.41 & 0.09 & 1.37 & $-0.02$ & 1.19 \\
  \bottomrule
  \end{tabular}%
  }
\end{table}

Our model predicts $\pi_{0.5}$'s task success with a root-mean-square (RMS) error of 5.7\% across the four tasks and three execution modes, with representative predictions shown in Figure~\ref{fig:prediction-summary}.
For this evaluation, we calibrate on zero-latency synchronous execution and all three modes at 100 and 200\,ms, then predict success at 50, 150, and 250\,ms.
Even with substantially less calibration data---one complete success curve plus one point from each of two other configurations---the median RMS error on the remaining measurements is 7.7\% across the evaluated calibration choices for $\pi_{0.5}$ (Table~\ref{tab:two-point} in Appendix~\ref{app:model}).
At 50\,ms latency, we also evaluate asynchronous execution with horizons of $H=2$, 3, and 4, which lie between those of continuous ($H=1$) and lazy ($H=5$) execution.
For these configurations, our model achieves RMS prediction errors of 3.3\% on \emph{set-down} and 2.8\% on \emph{release}.
Appendix~\ref{app:model} reports the full calibration and horizon results, along with separately calibrated predictions for other VLA policies.

We fit the model for each policy--task pair (Table~\ref{tab:sensitivities}) to show their sensitivity to staleness and pauses.
For $\pi_{0.5}$, \emph{grasping} is more sensitive to observation staleness, \emph{release} and \emph{articulated-object manipulation} to robot pauses, and \emph{set-down} to both.
MolmoAct2 follows the same pattern except on \emph{set-down}, where it is more sensitive to staleness.
GR00T N1.7 is an outlier: its fitted coefficients are negative for observation staleness on \emph{set-down} and robot idle time on \emph{grasping} (Table~\ref{tab:sensitivities}).
This behavior may partly reflect its sensitivity to initial target positions: changing these positions from LIBERO's original layouts reduces its zero-latency success to 52\% on \emph{release} and 56\% on \emph{set-down}, even with nearly static targets (Table~\ref{tab:policy-baseline}).
As a consequence, near-static \emph{set-down} targets accidentally yield even lower success than the fastest targets even at zero latency (Figures~\ref{fig:policy-summary} and~\ref{fig:gr00t-success-grid}).
These behaviors limit how well target-speed scaling captures its latency sensitivity.

\begin{table}[t]
  \centering
  \caption{Using measurements from System~A, our model predicts the best configurations for Systems~B--D. Each system favors a different execution mode and horizon.
  $^*$Calibration configuration.}
  \label{tab:deployment-systems}
  \footnotesize
  \setlength{\dashlinedash}{0.2pt}
  \setlength{\dashlinegap}{1.5pt}
  \setlength{\arrayrulewidth}{0.3pt}
  \setlength{\tabcolsep}{1.5pt}
  \renewcommand{\arraystretch}{1.15}
  \scalebox{0.8}{%
    \begin{tabular}{@{}M{10mm}!{\vrule width 0.5pt}L{24mm}!{\vrule width 0.5pt}R{12mm}!{\vrule width 0.5pt}R{16mm}!{\vrule width 0.5pt}M{16mm}!{\vrule width 0.5pt}R{18mm}@{}}
    \hline
    \rule[-2.5ex]{0pt}{6.5ex}\textbf{System} & \textbf{GPU / network} & \shortstack{\textbf{Latency}\\\textbf{(ms)}} & \shortstack{\textbf{Rate limit}\\\textbf{(req/s)}} & \shortstack{\textbf{Predicted}\\\textbf{Mode}, $\boldsymbol{H}$} & \shortstack{\textbf{Success (\%)}\\\textbf{Pred./Meas.}} \\
    \hline\hline
    A & Jetson AGX Thor & 72.8 & Uncapped & Cont., 2 & $84.5^{*}/84.5$ \\ \hdashline
    B & RTX 4090, WiFi~7 & 37.8 & 20.0 & Cont., 1 & $92.3/95.0$ \\ \hdashline
    C & H100, 5G & 42.9 & 5.0 & Async., 4 & $84.5/82.0$ \\ \hdashline
    D & \mbox{H100, slow cloud} & 217.9 & Uncapped & Lazy, 5 & $46.1/41.5$ \\
    \hline
    \end{tabular}%
  }
\end{table}

\subsection{System-Oriented Algorithm Space Exploration}
\label{sec:deployment-selection}

Achieving the highest task success rate on a given inference system requires choosing the right combination of execution mode and execution horizon.
The GPU, network, software runtime, and resource-sharing arrangement jointly determine the system's inference latency and available throughput per robot, so a configuration that performs well on one system may not be the best choice for another.
Exploring this configuration space experimentally is expensive because each candidate requires repeated task executions to estimate its success rate.
\textit{Our goal is therefore to reuse past deployment data to select configurations that maximize task success on new systems.}

We apply the prediction model from \S\ref{sec:eval-model} to four representative system deployment scenarios (Table~\ref{tab:deployment-systems}).
These comprise an onboard GPU with relatively high inference latency (System~A), a dedicated edge server with low latency (System~B), a shared server with low latency but limited inference throughput per robot (System~C), and a remote cloud server with high network latency (System~D).
GPU inference measurements (Table~\ref{tab:latency}) and network latency assumptions are detailed in Appendix~\ref{app:deployment}~\citep{su2026flashrt,jiang2026vlaperf}.
We calibrate the model for $\pi_{0.5}$ on \emph{set-down} using one complete success curve (continuous execution at 100\,ms) and one point from each of two additional curves (synchronous and continuous execution at 200\,ms).
We then predict success at a target speed of $0.075$\,m/s across feasible execution modes and horizons, selecting the configuration with the highest predicted success for each system.
The selected configurations match the measured best configurations on all four systems, with prediction errors ranging from 2.5\% to 4.6\% on Systems~B, C, and~D.
System~A selects continuous execution at 100\,ms, whose success curve is used for calibration.

\textbf{Different systems favor different execution modes and horizons, even for the same VLA policy and task} (Table~\ref{tab:deployment-systems}).
System~A favors continuous execution ($H=2$) to keep the robot moving while requesting inference as frequently as its onboard GPU allows.
System~B's lower latency supports continuous execution with a shorter horizon ($H=1$), providing fresher observations for precise placement.
System~C's limited per-robot throughput instead favors a longer asynchronous horizon ($H=4$), a mode between lazy and continuous asynchronous execution, that sustains robot motion within the request budget.
System~D's high network latency removes the staleness advantage of continuous execution, and the model favors lazy asynchronous execution ($H=5$) for this task.
Appendix~\ref{app:deployment} provides the complete success predictions and recommended configurations across inference latencies and per-robot throughput limits.

\section{Conclusion and Future Work}
\label{sec:conclusion}

\ours{} represents a foundational step toward understanding how VLA inference latency affects task success.
We identify the factors that shape latency sensitivity and compare execution modes and VLA policies across tasks and target speeds.
We further develop an empirical model that predicts task success and guides execution-mode selection for a given inference system.
We hope these findings will provide useful guidance for future robotics algorithm--system co-design solutions.

\label{body-end}
\bibliographystyle{mlsys2025}
\bibliography{references}
\section*{AI Usage Disclosure}
We used AI coding assistants (Claude Code and Codex) to write code, run data analysis, plot figures, search related work, check grammar, and draft some parts of the paper.
The authors directed and reviewed all AI-assisted work and take full responsibility for the paper's content.

\clearpage
\appendix
\onecolumn \pdfbookmark[1]{Appendix Contents}{appendix-contents}
\section*{Appendix Contents}

These appendices provide supporting details for the benchmark design, evaluation, and prediction model.

\begingroup
\setlength{\parindent}{0pt}
\newcommand{\appendixentry}[3]{%
  \par\medskip\noindent
  \hyperref[#1]{\textbf{Appendix~\ref*{#1}. #2}}\hfill
  \hyperref[#1]{p.~\pageref*{#1}}\par\nobreak
  #3\par\medskip}

\appendixentry{app:task-selection}{Task selection within LIBERO}{%
Selection criteria and how the other LIBERO tasks relate to the four chosen interactions.}

\appendixentry{app:protocol}{Evaluation and measurement details}{%
Task mappings, policy checkpoints, target paths and motion, episode design, statistical reporting, and definitions of attempts, tolerance zones, and hand speed.}

\appendixentry{app:full-curves}{Full success curves and recovery}{%
Complete results across target speeds, inference latencies, execution modes, and VLA policies.}

\appendixentry{app:model}{Model validation and calibration}{%
Held-out prediction accuracy and its noise floor, the reference curve, limited-data calibration, deployment predictions, and where speed scaling departs from the measurements.}

\appendixentry{app:dynamicvla}{DynamicVLA in Isaac Sim: a second simulator and policy}{%
Cross-simulator experiments on release and grasping with a policy trained on moving objects.}

\appendixentry{app:real-robot}{Real-robot experiments}{%
Experiments with MolmoAct2 on an SO-101 arm across grasping and release tasks, target speeds, inference latencies, and execution modes.}

\appendixentry{app:failures}{Failure stages}{%
Stage pass rates showing where latency costs success in each task.}

\appendixentry{app:related}{Additional related work}{%
Related approaches to asynchronous inference, policy execution speed, and latency compensation.}
\endgroup

\twocolumn

\section{Task selection within LIBERO}
\label{app:task-selection}

We selected the four tasks from LIBERO's four standard suites by three criteria: the evaluated checkpoints were trained on the task's suite and succeed in nearly all of its static episodes, each task adds an interaction the others lack, and moving the target lowers success enough for latency to matter.
LIBERO-90 fails the first criterion: neither $\pi_{0.5}$ nor GR00T N1.7 was trained on it, and from its official initial states they succeed in 28.3\% and 16.8\% of static episodes (GR00T with its LIBERO-10 checkpoint).
Within the four standard suites, most tasks repeat one of the four interactions or combine two subgoals.
All ten LIBERO-Object tasks put a grocery item into the basket (\emph{release}), and all ten LIBERO-Spatial tasks put the black bowl on the plate (\emph{set-down} when the plate moves, \emph{grasping} when the bowl's support moves).
Nine of the ten LIBERO-10 tasks combine two subgoals, such as putting two items into the basket or putting a bowl into a drawer and closing it.
LIBERO-Goal's tasks set objects down, put the cream cheese into a bowl (\emph{release}), open a drawer or turn on the stove (\emph{articulated-object manipulation}), or open a drawer and then place the bowl inside; the remaining task pushes the plate, so the robot itself moves the target.

\section{Evaluation and measurement details}
\label{app:protocol}

\begin{figure*}[tp]
  \centering
  \includegraphics[width=\linewidth]{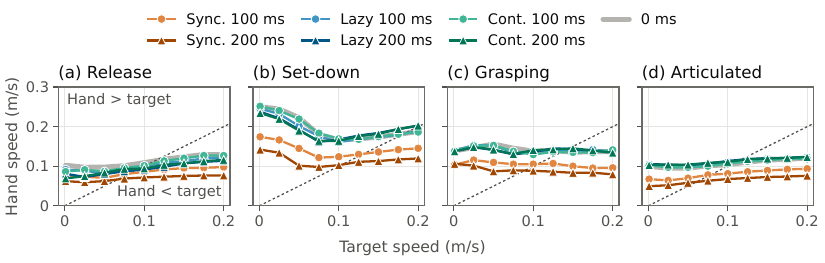}
  \caption{$\pi_{0.5}$ moves its hand at a nearly constant pace regardless of target speed, which asynchronous execution preserves and synchronous pauses reduce.
  Dotted lines mark hand speed equal to target speed.}
  \label{fig:hand-speed}
\end{figure*}

Targets follow fixed closed paths while the robot, the cameras, the language instructions, the success checks, and MuJoCo's contact model stay as in LIBERO; this section also defines the attempts, tolerance zones, and hand speed behind Figures~\ref{fig:tolerance} and~\ref{fig:hand-speed}.
\emph{Release} uses LIBERO-Object task~1, whose basket containment region is 12.2\,cm wide; \emph{set-down} and \emph{grasping} use LIBERO-Spatial task~9, whose success check requires the bowl's center within 3\,cm of the plate's; and \emph{articulated-object manipulation} uses LIBERO-Goal task~0.
GR00T uses suite-specific checkpoints~\citep{nvidia2026gr00tlibero}, and MolmoAct2 uses one checkpoint for all suites~\citep{fang2026molmoact2}.

Each target moves at constant speed along one closed cubic spline through 12 waypoints on an ellipse; the paths span about $22\times14$\,cm for \emph{release}, $21\times14$\,cm for \emph{set-down}, $11\times13$\,cm for \emph{grasping}, and $7\times10$\,cm for \emph{articulated-object manipulation}.
We evaluate every task at $0.001$\,m/s and from $0.025$ to $0.2$\,m/s in $0.025$\,m/s steps.
The grid stops at $0.2$\,m/s, past the speed at which $\pi_{0.5}$'s zero-latency success falls to 50\% on every task ($0.04$--$0.19$\,m/s); at $0.2$\,m/s it succeeds in 5.0--46.0\% of episodes, against 93.0--98.5\% near static.
We keep the background objects where LIBERO's original layouts place them, since perturbing initial object states can sharply reduce policy success~\citep{zhou2025liberopro}; in \emph{set-down} the stationary cabinet holding the bowl starts at the same position in every layout, and in \emph{grasping} the bowl starts at the center of its moving support.
Each condition's 200 episodes cover LIBERO's 50 official initial states, each with two starting points on the path and both travel directions.

\textit{Task deadlines.}
Each LIBERO episode has a budget of 600 executed policy actions, increased from 220--300 in standard static LIBERO to allow more time to pursue moving targets.
An episode is considered a failure if the task remains incomplete when this budget is exhausted.
Stationary actions inserted during inference waits do not consume this budget: we extend the deadline by one control step for each such idle step while the target continues moving.
This preserves the same executed-action budget across inference latencies and execution modes.

\phantomsection
\label{app:motion}
A driver writes each target's planar path at every physics substep.
The basket and the plate are free bodies whose planar position and velocity are written while their height and rotation stay physical; the bowl's support moves on added planar slide joints whose position and velocity follow the path; and the drawer's cabinet, a jointless LIBERO fixture, is moved by position alone while the drawer keeps its damped slide joint.
No gravity, mass, friction, or solver setting changes, so MuJoCo resolves every contact with its unmodified contact model.

Following OpenPI's protocol~\citep{openpi2025libero}, each episode begins with 10 warm-up steps in which the robot and the target stay still.
In asynchronous modes, the robot waits only for the first chunk and executes it from its first action.
Synchronous and lazy execution coincide at zero latency; the policy comparison in Appendix~\ref{app:full-curves} pools their zero-latency runs where both exist, while the success curves and Table~\ref{tab:policy-baseline} use synchronous runs.
Intervals are 95\% Wilson confidence intervals.

A tolerance zone is the range of horizontal offsets from the target center within which at least half of the attempts hit.
We pool attempts from all speeds, latencies, and modes, retries included, into 0.5\,cm bins of offset; a bin with at least 30 attempts belongs to the zone if at least half of them hit.

Hand speed is the hand's path length per second while it closes in on the target, over every qualifying interval after warm-up, retries and inference waits included.
For \emph{release} and \emph{set-down}, an interval qualifies while both fingers hold the object within 15\,cm horizontally of the receptacle or surface and the hand is at or above the receptacle's rim or the surface; for \emph{grasping} and \emph{articulated-object manipulation}, it qualifies while the empty hand is within 10\,cm of the object or handle without touching it.
Figure~\ref{fig:hand-speed} plots the median episode hand speed at each target speed; the 34\% reduction in Section~\ref{sec:eval-tasks} compares the medians of these values over target speeds.

\phantomsection
\label{app:events}
Attempts are detected from the MuJoCo state traces, recorded once per 50\,ms control step, where two bodies are in contact when their contact distance is $\leq0$.
A \emph{confirmed pickup} starts the first uninterrupted run of at least three samples with both fingers in contact with the object in which the object is at least once 2\,cm above its post-warm-up height.
A \emph{release attempt} begins a run of at least two samples without finger contact, after contact by both fingers within the preceding three samples, with the object lifted at least 2\,cm and the gripper commanded open; its offset is measured at the last contact sample, and it hits if the episode succeeds without later contact.
A \emph{set-down attempt} begins at each onset of object--surface contact, with gaps shorter than three samples merged, and hits if the task completes during that contact.
A \emph{grasp attempt} is the first sample with the lowest fingertip below the object's rim after at least three samples above it, with the hand within 15\,cm horizontally of the object, no later than the start of the first confirmed pickup or the first sample with the object more than 5\,cm below its post-warm-up height; it hits if a confirmed pickup starts at or after it, before the next attempt, and within 40 samples.

\FloatBarrier
\section{Full success curves and recovery}
\label{app:full-curves}

\begin{figure*}[tp]
  \centering
  \includegraphics[width=\linewidth]{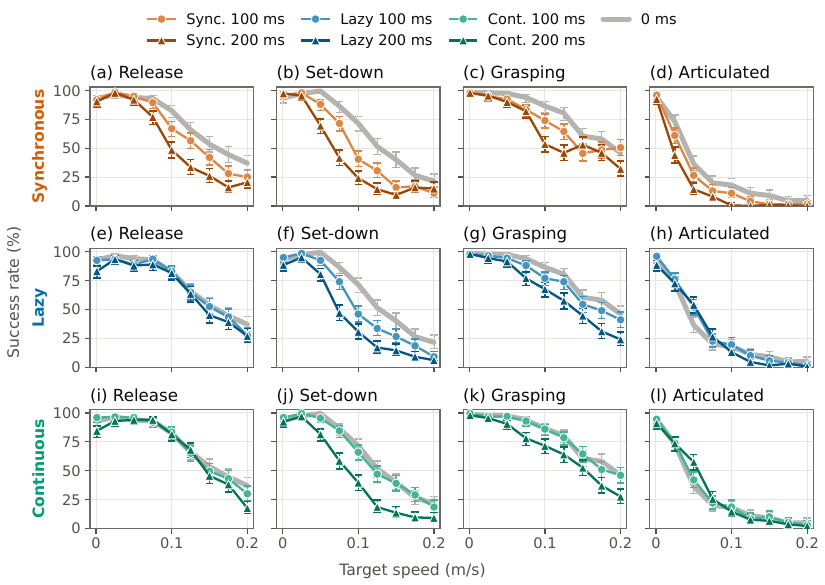}
  \caption{At 100 and 200\,ms latencies, lazy execution eliminates most synchronous losses on release and articulated-object manipulation, whereas at 200\,ms set-down loses success on moving targets in all three modes.
  Curves show $\pi_{0.5}$'s measured success at the calibration latencies.}
  \label{fig:app-success-grid}
\end{figure*}

\begin{figure*}[tp]
  \centering
  \includegraphics[width=\linewidth]{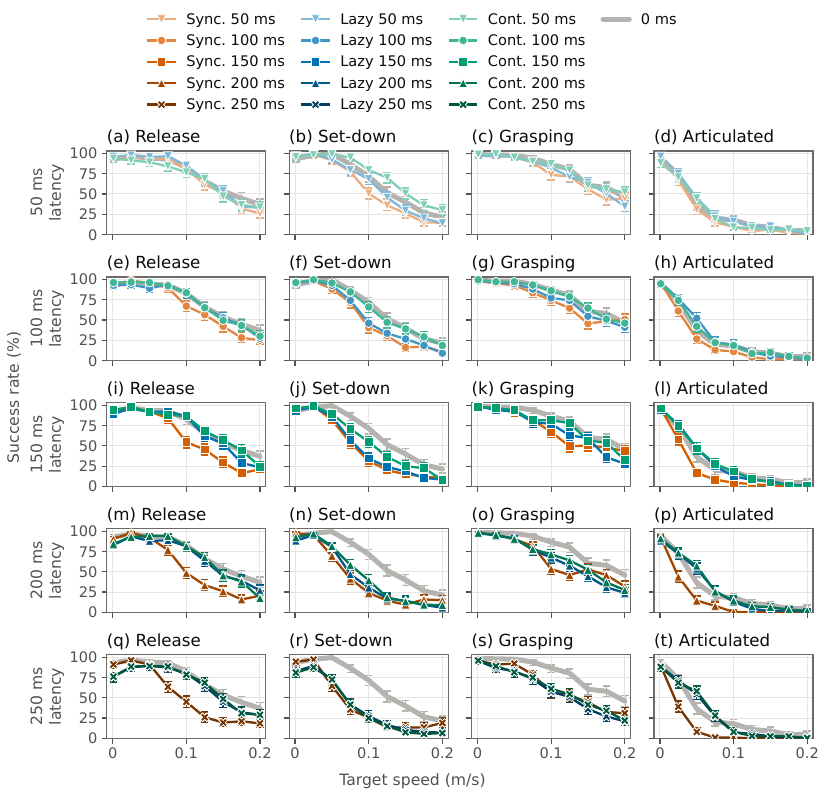}
  \caption{Continuous execution outperforms zero-latency execution on set-down for $\pi_{0.5}$ at 50\,ms before converging toward lazy execution as latency approaches lazy execution's horizon, while 250\,ms asynchronous execution loses success even near static.}
  \label{fig:app-mode-comparison}
\end{figure*}

\begin{figure*}[tp]
  \centering
  \includegraphics[width=\linewidth]{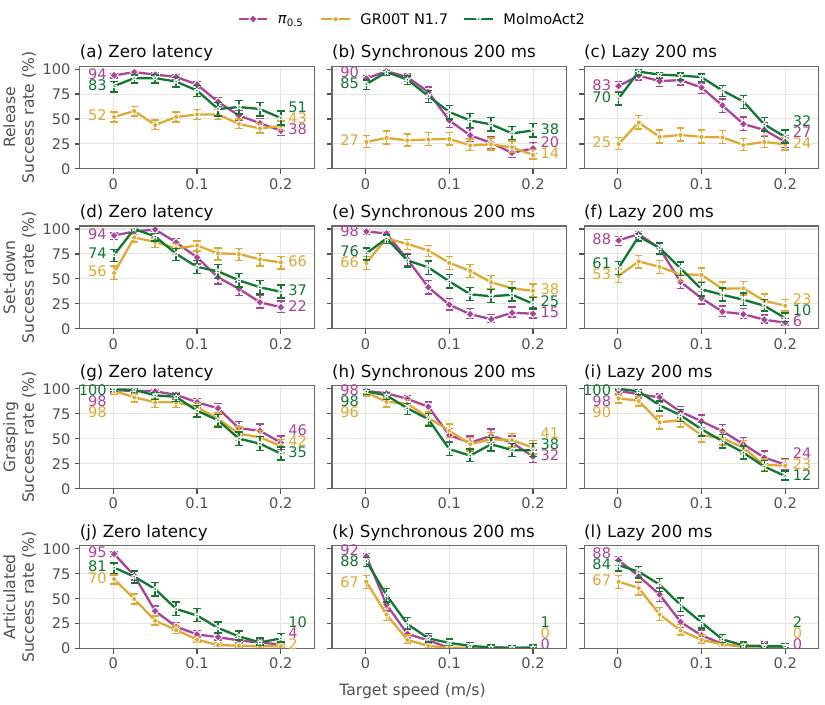}
  \caption{Policy rankings invert across tasks and speeds: GR00T N1.7 trails near static on three tasks but surpasses $\pi_{0.5}$ on fast set-down targets, while MolmoAct2 leads on moving articulated targets.}
  \label{fig:policy}
\end{figure*}

\begin{figure*}[tp]
  \centering
  \includegraphics[width=\linewidth]{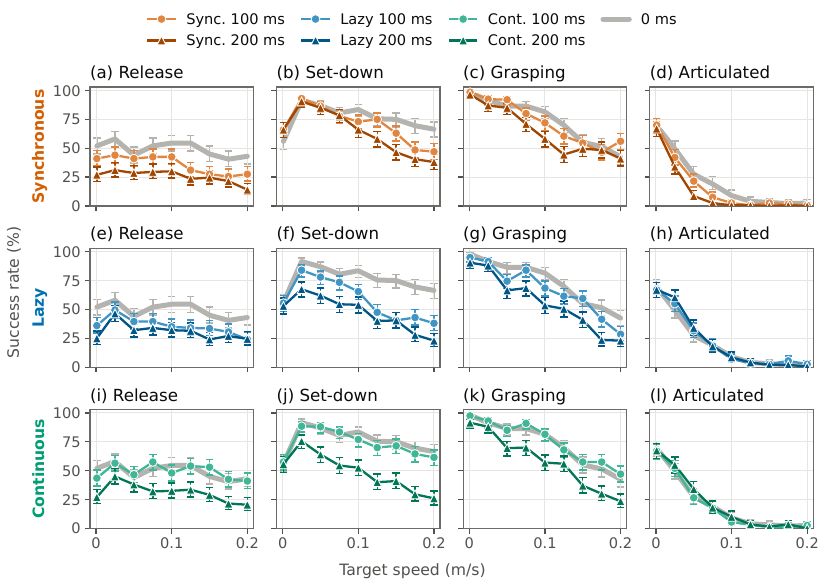}
  \caption{On articulated-object manipulation at $0.05$\,m/s, lazy execution at 200\,ms raises GR00T N1.7's success from 8.5\% under synchronous execution to 34.0\% (27.5\% at zero latency), but on release at $0.1$\,m/s only from 30.0\% to 32.0\%; on grasping at 200\,ms, synchronous execution beats lazy execution at six of eight moving-target speeds.}
  \label{fig:gr00t-success-grid}
\end{figure*}

\begin{figure*}[tp]
  \centering
  \includegraphics[width=\linewidth]{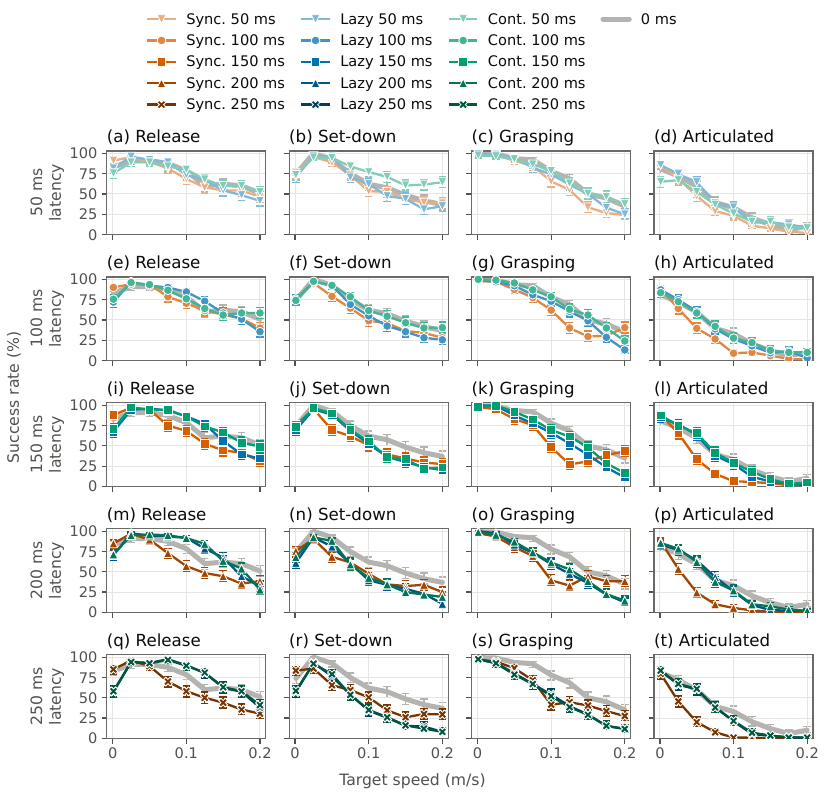}
  \caption{On MolmoAct2, continuous execution at 50\,ms exceeds zero-latency success on set-down from $0.075$\,m/s, by 9.0--27.5 points, and zero-latency success is lower at $0.001$\,m/s than at $0.025$\,m/s on release and set-down.}
  \label{fig:molmoact2-mode-comparison}
\end{figure*}

Figures~\ref{fig:app-success-grid}, \ref{fig:app-mode-comparison}, \ref{fig:policy}, \ref{fig:gr00t-success-grid}, and~\ref{fig:molmoact2-mode-comparison} give every tested speed, latency, execution mode, and policy behind Section~\ref{sec:evaluation}.
On MolmoAct2, zero-latency success is higher at $0.025$\,m/s than at $0.001$\,m/s on \emph{release} (91.0\% vs.\ 83.0\%) and \emph{set-down} (100\% vs.\ 74.0\%; Figure~\ref{fig:molmoact2-mode-comparison}).
In its near-static set-down failures, MolmoAct2 tends to leave the bowl on the plate's rim and then stop, whereas a plate moving at $0.025$\,m/s drags the bowl onto itself.

\FloatBarrier

\FloatBarrier

\FloatBarrier

\FloatBarrier

\FloatBarrier
\section{Model validation and calibration}
\label{app:model}

\begin{figure*}[tp]
  \centering
  \includegraphics[width=\linewidth]{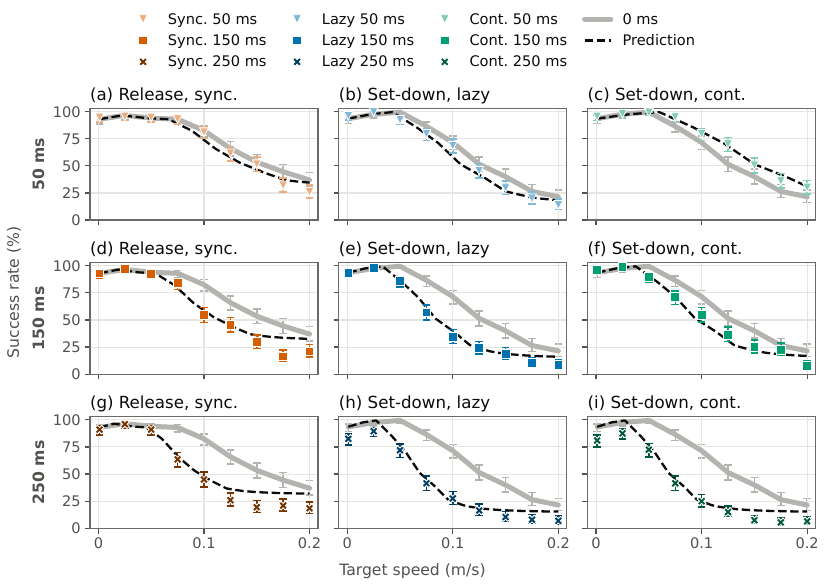}
  \caption{Speed-scaling predictions fitted on 0, 100, and 200\,ms runs track $\pi_{0.5}$'s held-out 50, 150, and 250\,ms curves in three representative settings; over every task and mode, the held-out error grows with latency (4.2, 5.5, and 7.0 points RMS).
  Dashed black curves show model predictions beside measured markers.}
  \label{fig:prediction-examples}
\end{figure*}

\begin{figure*}[tp]
  \centering
  \includegraphics[width=\linewidth]{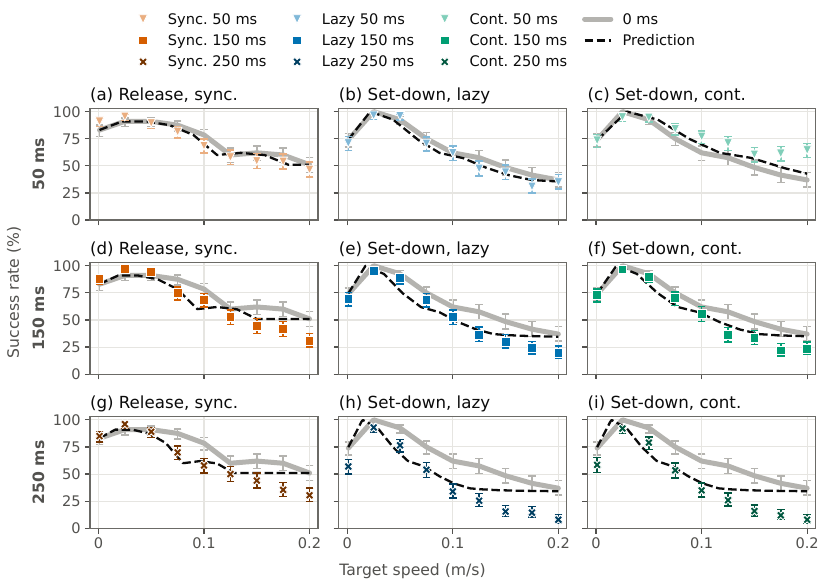}
  \caption{Held-out predictions on MolmoAct2 capture speed scaling across latencies and carry the zero-latency dip of near-static set-down, with larger residuals at 250\,ms.
  Dashed black curves show predictions from 0, 100, and 200\,ms fits beside measured markers.}
  \label{fig:molmoact2-prediction-examples}
\end{figure*}

Fitted on 0, 100, and 200\,ms runs, the model predicts the held-out 50, 150, and 250\,ms curves with an RMS error of 5.7 points for $\pi_{0.5}$ and 7.8 points for MolmoAct2 (Table~\ref{tab:blind}, Figures~\ref{fig:prediction-examples} and~\ref{fig:molmoact2-prediction-examples}).
Sampling noise alone would make an exactly right model miss these measurements by 4.0 points RMS for $\pi_{0.5}$ (3.9 for MolmoAct2), combining the held-out points' own binomial noise, 2.6 points (2.8), with that of the measured reference curve each prediction reads.

Each prediction reads the reference curve, the measured success curve of the reference configuration, at the target speed scaled by $\lambda$ from Equation~\ref{eq:model}.
Between measured target speeds the curve interpolates the measured success rates linearly, with no monotonicity imposed; below the slowest measured speed it keeps that speed's success rate; above the fastest it follows a log-logistic curve fitted by maximum likelihood to the reference's own points and scaled to meet the measured success at that speed.
Apart from this tail, which depends on the reference curve alone, the model's only free parameters are $a$ and $b$, fitted by binomial maximum likelihood on the calibration conditions.
For $\pi_{0.5}$, the model also predicts asynchronous execution at 50\,ms with intermediate horizons of $H=2$, 3, and 4, which no calibration includes, with RMS errors of 3.3 points on \emph{set-down} and 2.8 points on \emph{release}.

\begin{table}[htbp]
  \centering
  \small
  \caption{Calibrating Equation~\ref{eq:model} on 0, 100, and 200\,ms runs predicts held-out 50, 150, and 250\,ms curves with an RMS error of 5.7 percentage points on $\pi_{0.5}$ and 7.8 points on MolmoAct2.}
  \label{tab:blind}
  \begin{tabular}{@{}lccccc@{}}
  \toprule
  RMS error (points) & Release & Set-down & Grasping & Articulated & All \\
  \midrule
  \multicolumn{6}{@{}l}{\textbf{$\boldsymbol{\pi_{0.5}}$}} \\
    \quad Synchronous & 7.1 & 2.8 & 5.5 & 3.8 & 5.1 \\
  \quad Lazy & 6.3 & 5.3 & 5.7 & 5.2 & 5.6 \\
  \quad Continuous & 6.5 & 5.7 & 6.7 & 6.1 & 6.3 \\
  \quad Every mode & 6.7 & 4.8 & 6.0 & 5.1 & 5.7 \\
  Point sampling noise & 2.7 & 2.6 & 2.8 & 2.2 & 2.6 \\
  \midrule
  \multicolumn{6}{@{}l}{\textbf{MolmoAct2}} \\
    \quad Synchronous & 8.0 & 5.0 & 9.8 & 3.1 & 7.0 \\
  \quad Lazy & 11.2 & 10.2 & 4.2 & 4.9 & 8.2 \\
  \quad Continuous & 8.3 & 11.3 & 5.6 & 6.0 & 8.1 \\
  \quad Every mode & 9.3 & 9.2 & 7.0 & 4.8 & 7.8 \\
  Point sampling noise & 2.9 & 3.0 & 2.8 & 2.5 & 2.8 \\
  \bottomrule
  \end{tabular}
\end{table}

\FloatBarrier
\phantomsection
\label{app:calibration}
Our model can be calibrated without zero-latency data from one complete curve plus one point on each of two other curves.
Because $S_Y(v)\approx S_X(v\,\lambda_Y/\lambda_X)$ for any two settings $X$ and $Y$, the complete curve $X$, read as measured, fixes the curve shape, and the two points constrain the ratios $\lambda_Y/\lambda_X$, from which $a$ and $b$ are fitted by maximizing the binomial likelihood times a weak Gaussian prior over a bounded grid.
One of the three curves must be synchronous at a nonzero latency, because only it separates idle time from staleness.

We score every such calibration on the 0--250\,ms grid for $\pi_{0.5}$ and MolmoAct2: each complete curve at 50 to 250\,ms in any mode, with one point in each of the other two execution modes, both at the grid speed nearest that curve's midpoint speed divided by 1.4, where its measured success first falls halfway between its highest and lowest value (Table~\ref{tab:two-point}).
On $\pi_{0.5}$, the median RMS error over all remaining points is 7.7 points, 80\% of calibrations fall within 4.5--14.4 points, and the unseen zero-latency curve is recovered within a median of 6.8 points.
MolmoAct2's median is 11.1 points.
GR00T N1.7, measured only at 0, 100, and 200\,ms, has a median of 11.6 points.
For $\pi_{0.5}$, measuring both points on the sloped part of the curve, near the midpoint speed divided by 1.4, matters: over every pair of grid speeds, its median error rises to 9.4 points.

\begin{table}[htbp]
  \centering
  \caption{One complete curve, with one point on each of two other curves, predicts the remaining curves, including the unmeasured zero-latency curve.}
  \label{tab:two-point}
  \footnotesize
  \begin{tabular}{@{}lccccc@{}}
\toprule
RMS error (points) & Release & Set-down & Grasping & Articulated & All \\
\midrule
\multicolumn{6}{@{}l}{\textbf{$\boldsymbol{\pi_{0.5}}$}: one complete curve and two points, no zero-latency data} \\
\quad Median (80\% of calibrations) & 8.5 (5.4--18.4) & 6.8 (4.2--15.0) & 9.1 (6.2--13.0) & 5.4 (4.2--10.9) & 7.7 (4.5--14.4) \\
\quad Its zero-latency curve, median & 5.9 & 7.2 & 8.0 & 5.8 & 6.8 \\
\quad Fit at 0, 100, 200\,ms; held-out latencies & 6.7 & 4.8 & 6.0 & 5.1 & 5.7 \\
\midrule
\multicolumn{6}{@{}l}{\textbf{MolmoAct2}: one complete curve and two points, no zero-latency data} \\
\quad Median (80\% of calibrations) & 12.3 (8.6--19.0) & 12.2 (8.1--24.3) & 10.8 (8.2--14.0) & 7.4 (4.8--19.2) & 11.1 (6.5--19.4) \\
\quad Its zero-latency curve, median & 10.9 & 10.3 & 8.6 & 6.3 & 8.6 \\
\quad Fit at 0, 100, 200\,ms; held-out latencies & 9.3 & 9.2 & 7.0 & 4.8 & 7.8 \\
\bottomrule
\end{tabular}
\end{table}

\phantomsection
\label{app:deployment}
Table~\ref{tab:latency} lists the 99th-percentile latency of a single in-process request at bf16 with openpi, run in PyTorch, and with FlashRT~\citep{su2026flashrt}.
For networked deployments, we add approximately 5\,ms of network latency over WiFi~7, 25\,ms over 5G, and 200\,ms over a wired link to a slow cloud region, following the estimates in VLA-Perf~\citep{jiang2026vlaperf}.
System~A's latency is instead the median that VLA-HWBench reports for $\pi_{0.5}$ under FlashRT at 16-bit precision on an NVIDIA Jetson AGX Thor, with our workload of 10 actions and two camera images~\citep{vlahwbench2026}.

\begin{table}[tbp]
  \centering
  \caption{FlashRT cuts $\pi_{0.5}$'s 99th-percentile inference latency (ms) by 46--73\% relative to openpi on every GPU.}
  \label{tab:latency}
  \begin{tabular}{@{}lrr@{}}
  \toprule
  GPU & openpi & FlashRT \\
  \midrule
  RTX 3090 & 131.6 & 66.3 \\
  RTX 4090 & 60.6 & 32.8 \\
  RTX 5090 & 56.8 & 27.4 \\
  H100 & 66.3 & 17.9 \\
  \bottomrule
  \end{tabular}
\end{table}

\begin{figure}[tp]
  \centering
  \includegraphics[width=\linewidth]{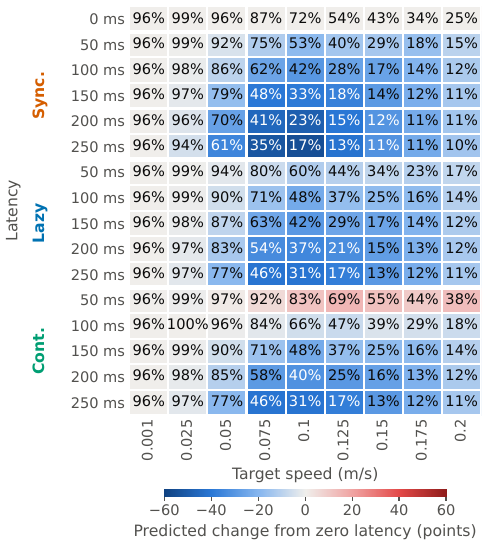}
  \caption{On set-down, the model predicts the highest $\pi_{0.5}$ success for continuous execution at nearly every moving-target speed up to 200\,ms latency and for lazy and continuous execution, tied, at 250\,ms; synchronous execution does not lead at any speed from $0.025$ to $0.2$\,m/s.
  Cells show predicted success across target speeds and latencies from calibrating on one complete curve (continuous execution at 100\,ms) plus one point on each of two other curves (synchronous and continuous execution at 200\,ms).}
  \label{fig:deployment-modes}
\end{figure}

From the calibration of Section~\ref{sec:deployment-selection}, the continuous 100\,ms curve plus synchronous and continuous points at 200\,ms, the model predicts $\pi_{0.5}$'s set-down success at every speed, latency, and mode (Figure~\ref{fig:deployment-modes}); FlashRT sets only the latency, and every success rate comes from our openpi JAX runs.
Each latency takes effect at the next whole 50\,ms control step: 37.8 and 42.9\,ms act as 50\,ms, 72.8\,ms as 100\,ms, and 217.9\,ms as 250\,ms.
Across the 133 measured points outside the calibration, the model errs by 4.2 points RMS against 2.6 points of sampling noise in the measured points (Table~\ref{tab:deployment-error}).
Its predicted best mode matches the measured best in 36 of 45 latency-by-speed cells, counting ties, including every latency at $0.075$\,m/s; at 250\,ms, lazy and continuous execution run the same schedule and are predicted equal.
All nine misses fall at 150--250\,ms; in each, the model ranks its choice less than 3 points above the measured best mode, and the measured best mode beats the choice by at most 7.5 points, except at 250\,ms and $0.2$\,m/s (12.0).

\begin{table}[htbp]
  \centering
  \caption{The model predicts $\pi_{0.5}$'s set-down success in every execution mode within about twice the sampling noise (2.6 percentage points RMS), slightly overestimating asynchronous execution.}
  \label{tab:deployment-error}
  \begin{tabular}{@{}lrrr@{}}
  \toprule
  Mode & Points & RMS error & Mean error \\
  \midrule
  Synchronous & 53 & 3.4 & $-0.6$ \\
  Lazy & 45 & 4.5 & $+2.6$ \\
  Continuous & 35 & 5.0 & $+2.8$ \\
  \midrule
  All & 133 & 4.2 & $+1.4$ \\
  \bottomrule
  \end{tabular}
\end{table}

\begin{figure}[tp]
  \centering
  \includegraphics[width=\linewidth]{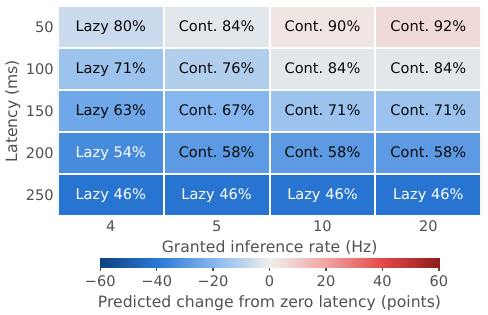}
  \caption{On set-down at $0.075$\,m/s, continuous execution gives $\pi_{0.5}$ the highest predicted success up to 200\,ms at 5\,Hz or more, and lazy execution at 4\,Hz and at 250\,ms.
  Under a rate below one inference per latency, continuous execution requests once per granted period and switches to each chunk on arrival.}
  \label{fig:deployment-rate}
\end{figure}

Figure~\ref{fig:deployment-rate} gives, from the same calibration, the best schedule each granted inference rate allows at each latency.

\phantomsection
\label{app:misses}
Speed scaling departs from the measurements in two places.
First, on $\pi_{0.5}$'s fastest targets, synchronous execution keeps more success than both asynchronous modes: on \emph{grasping} at $0.2$\,m/s at every latency from 100 to 250\,ms (200\,ms: 32.0\% against 24.0\% lazy and 27.5\% continuous) and on \emph{set-down} at $0.175$--$0.2$\,m/s at 200 and 250\,ms (250\,ms, $0.2$\,m/s: 19.0\% against 7.0\% and 6.5\%; Figure~\ref{fig:app-mode-comparison}).
The model gives synchronous execution lazy execution's staleness plus idle time, so on these falling parts of the reference curve it predicts synchronous execution 0.2--1.6 points below lazy execution.
Second, at 250\,ms, lazy and continuous execution lose success near static ($0.001$\,m/s) on \emph{release} (93.0\% to 76.0\% in both modes) and \emph{set-down} (93.5\% to 82.5\% and 81.0\%), where after the first chunk they execute actions 5--9 of each chunk, while synchronous execution, executing actions 0--4, stays within two points (91.0\%, 94.5\%); the model moves these near-static predictions by less than 0.2 points.
Points read from the extrapolated reference curve beyond $0.2$\,m/s, about a fifth of the held-out points, carry 33\% of $\pi_{0.5}$'s held-out squared error and 37\% of MolmoAct2's, whose held-out error gathers at the three fastest speeds ($0.15$--$0.2$\,m/s: 58\%, against 44\% for $\pi_{0.5}$).

\FloatBarrier
\section{DynamicVLA in Isaac Sim: a second simulator and policy}
\label{app:dynamicvla}

\begin{figure*}[tp]
  \centering
  \includegraphics[width=\linewidth]{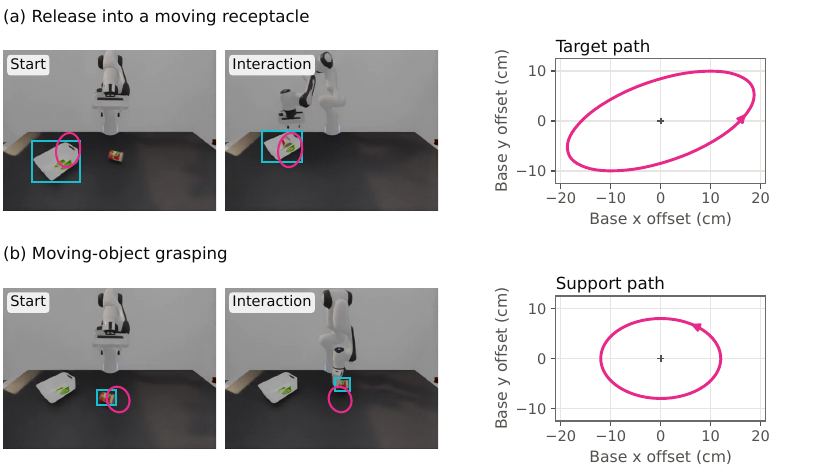}
  \caption{Our Isaac Sim verification replicates release and grasping with DynamicVLA along closed trajectories in robot-base coordinates.
  A cyan box marks the moving object and pink traces its path; an unrendered rail carrier moves the grasping can.}
  \label{fig:dynamicvla-tasks}
\end{figure*}

We test the official DOM checkpoint of DynamicVLA~\citep{xie2026dynamicvla} in Isaac Sim on \emph{release} and \emph{grasping} at a 25\,Hz control rate with an execution horizon of $H=14$ and a chunk size of $K=20$.
A driver writes the planar position and velocity of each moving body before every control step: the box in \emph{release} and, in \emph{grasping}, an unrendered rail carrier that moves the can through contact alone (Figure~\ref{fig:dynamicvla-tasks}).
We run synchronous execution at 0, 120, and 240\,ms and lazy and continuous execution at 120 and 240\,ms, where 240\,ms ($L=6$) is the longest latency satisfying $K\geq H+L$.
Each setting runs 200 episodes at each target speed.
Each episode has a budget of 750 executed policy actions, increased from DOM's native 300 to allow more time to pursue moving targets, and DOM's drop checks end it early as a failure if the can or the box falls off the table.

\begin{figure}[t]
  \centering
  \includegraphics[width=\linewidth]{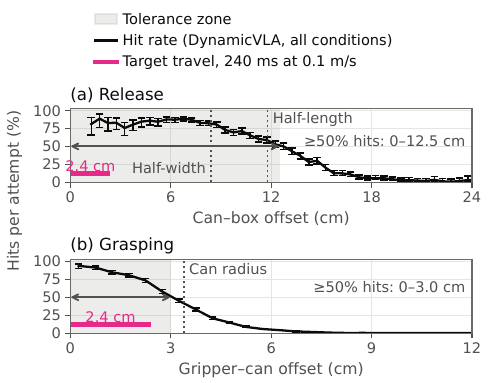}
  \caption{Hit rate of DynamicVLA's attempts against their horizontal offset from the target center when the can leaves the gripper (\emph{release}) or the fingertips descend below the can's top (\emph{grasping}).
  \emph{Grasping} requires greater precision than \emph{release}.}
  \label{fig:dynamicvla-tolerance}
\end{figure}

\begin{figure*}[tp]
  \begin{minipage}[t]{.5\linewidth}
    \vspace{0pt}
    \includegraphics[width=\linewidth]{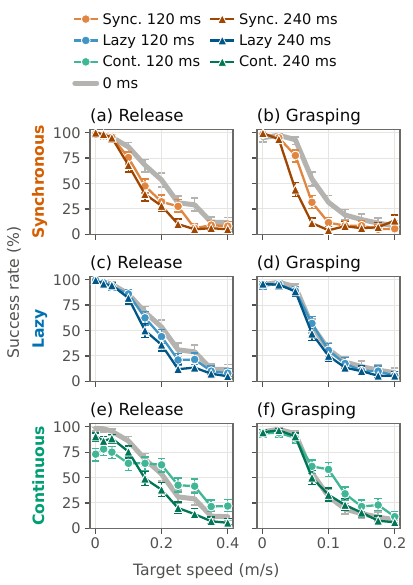}
    \caption{On DynamicVLA in Isaac Sim, near-static grasping stays at 93.5\% or higher in every setting, synchronous latency moves grasping success to lower speeds, and asynchronous grasping never falls more than 7.5 points below its zero-latency curve, while 120\,ms continuous execution exceeds it from 0.075\,m/s.}
    \label{fig:dynamicvla-success-grid}
  \end{minipage}\hfill
  \begin{minipage}[t]{.5\linewidth}
    \vspace{0pt}
    \includegraphics[width=\linewidth]{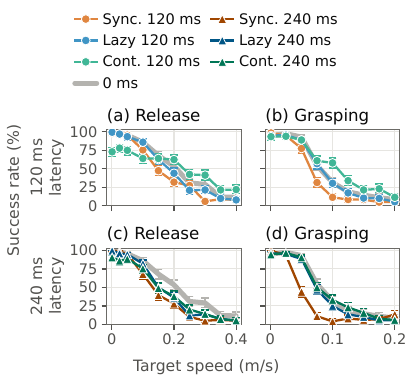}
    \caption{Lazy execution keeps DynamicVLA's grasping within 7.5 points of its zero-latency curve at 240\,ms, while 120\,ms continuous execution raises success on fast targets and lowers it on slow release targets.}
    \label{fig:dynamicvla-mode-comparison}
  \end{minipage}
\end{figure*}

\textbf{Finding~1 holds.} Inference latency substantially reduces task success on moving targets, to an extent that depends on the task and target speed.
Synchronous execution at 240\,ms does not lower near-static success on either task (release 98.5\% to 100\%, grasping 95.0\% to 99.5\%), but it lowers the target speed at which success falls to 50\% from 0.21 to 0.13\,m/s on \emph{release} and from 0.080 to 0.047\,m/s on \emph{grasping} (Figure~\ref{fig:dynamicvla-success-grid}a,b).

\textbf{Finding~2 holds.} A task's latency sensitivity depends on its requirements for action precision and continuous motion.
Figure~\ref{fig:dynamicvla-tolerance} compares the precision the two tasks require: \emph{grasping} requires closing the fingers within 3\,cm of the can's center, whereas \emph{release} tolerates drops within 12.5\,cm of the box's center.
At a target speed of 0.1\,m/s, 240\,ms of additional observation staleness allows the target to move 2.4\,cm, four-fifths of \emph{grasping}'s tolerance zone.

\textbf{Finding~3 holds, except on slow release targets.} Asynchronous execution recovers task success, and continuous asynchronous execution recovers the most.
At 240\,ms, lazy execution keeps \emph{grasping} within 7.5 points of zero latency at every speed, but raises the 50\%-success speed of \emph{release} only from synchronous execution's 0.13 to 0.15\,m/s, against 0.21\,m/s at zero latency (Figure~\ref{fig:dynamicvla-mode-comparison}c,d).
Continuous execution at 120\,ms acts on fresher observations than zero-latency synchronous execution (160\,ms against 260\,ms of average staleness) and raises success on fast targets in both tasks (Figure~\ref{fig:dynamicvla-mode-comparison}a,b).
The exception is \emph{release} up to 0.1\,m/s, where continuous execution falls below lazy execution at both latencies: near static, it succeeds in 73.0\% of episodes at 120\,ms and 90.5\% at 240\,ms, against 99.5\% for lazy execution (Figure~\ref{fig:dynamicvla-mode-comparison}a,c).
At 120\,ms, it drops the can early: in 36 of its 54 near-static failures, the gripper is commanded open while the can is still more than 3\,cm up and more than 20\,cm from the box.

The model of Section~\ref{sec:eval-model} fits DynamicVLA's \emph{grasping} within about twice the sampling noise and its \emph{release} less closely, mostly where continuous execution drops the can early.
Fitted on every curve at nonzero latency, Equation~\ref{eq:model} gives \emph{grasping} $a=0.16$ (relative to DynamicVLA's 260\,ms reference staleness) and $b=1.37$, with an in-sample RMS error of 5.0 points against 2.4 points of sampling noise in the measured points.
Pauses are thus the main cost of latency for DynamicVLA's \emph{grasping}, unlike for $\pi_{0.5}$'s, whose $b$ is 0.07 (Table~\ref{tab:sensitivities}).
On \emph{release}, the fit gives $a=0.38$ and $b=0.38$, with an RMS error of 7.9 points against 2.6, most of it from continuous execution at 120\,ms on targets up to 0.1\,m/s, where measured success falls 20--26 points below the fit.

DynamicVLA also moves differently from $\pi_{0.5}$: it swings its hand wide even toward a near-static can.
In near-static episodes, its hand path to a 2\,cm lift of the can is more than 2.5 times the straight-line distance in 86.5--100\% of lifting episodes in every setting except continuous execution at 120\,ms (28\% on \emph{release}, 24\% on \emph{grasping}), swinging a median 40\,cm to the side of the can; $\pi_{0.5}$'s hand path does so in at most 0.5\% of near-static release episodes.

\FloatBarrier
\section{Real-robot experiments}
\label{app:real-robot}

We test the official checkpoint of MolmoAct2 for the \mbox{SO-100} and \mbox{SO-101} arms~\citep{fang2026molmoact2} on an \mbox{SO-101} arm~\citep{therobotstudio2025soarm100} with a front camera and a wrist camera at a 30\,Hz control rate with an execution horizon of $H=16$ and its default chunk size of $K=30$.
We do not test $\pi_{0.5}$ or GR00T N1.7 on the robot because neither has an official checkpoint for this arm, and, as in simulation (\S\ref{sec:vla-policies}), we evaluate only official public checkpoints rather than policies we fine-tune ourselves.

\begin{figure*}[t]
  \centering
  \includegraphics{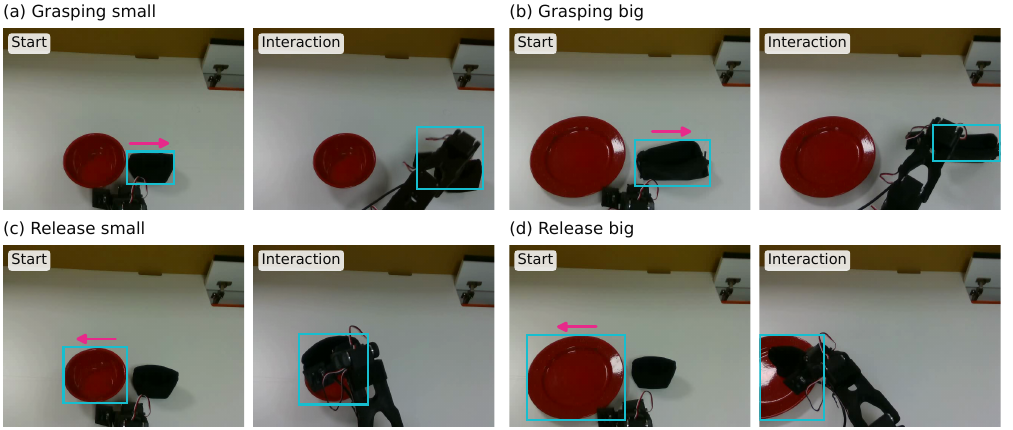}
  \caption{The four real-robot tasks pair grasping and release with a small and a big target, shown from the front camera at the start of a rollout and as the robot interacts with the moving target.
  Each moving target, highlighted by a cyan box, travels once along a straight path in the direction of the pink arrow.}
  \label{fig:real-robot-tasks}
\end{figure*}

\begin{figure*}[t]
  \centering
  \includegraphics{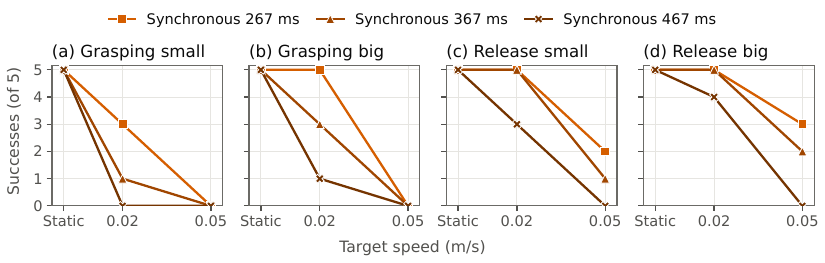}
  \caption{Task success on the SO-101 arm across target speeds and inference latencies using MolmoAct2 with synchronous execution.
  Static targets succeed in every rollout, while moving targets lose success as latency increases.}
  \label{fig:real-robot-sync}
  \vspace{1em}
  \includegraphics{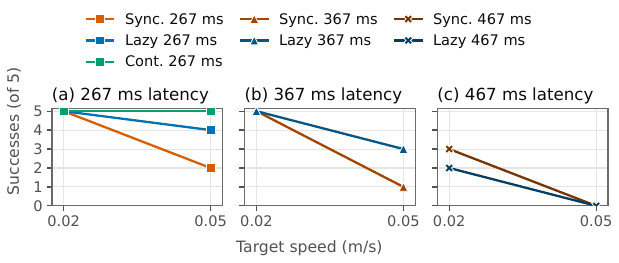}
  \caption{Task success on \emph{release small} using MolmoAct2 under synchronous and asynchronous execution at each inference latency.
  Continuous execution, and lazy execution at shorter latencies, recover success on the faster target.}
  \label{fig:real-robot-modes}
\end{figure*}

We select four tasks that pair two interaction types with a small and a big target: a $10\times6$\,cm mini tote bag 7\,cm tall and a $20\times7$\,cm case 8\,cm tall as the objects to grasp, and a bowl 15\,cm in diameter and 5\,cm tall and a plate 25\,cm in diameter and 2\,cm tall as the receptacles.
In \textbf{(a) grasping small}, the robot picks up the moving tote bag and places it in a stationary bowl.
In \textbf{(b) grasping big}, the robot picks up the moving case and places it on a stationary plate.
In \textbf{(c) release small} and \textbf{(d) release big}, the robot picks up the stationary tote bag and drops it into the moving bowl or onto the moving plate (Figure~\ref{fig:real-robot-tasks}).

Every moving target travels once, in one direction, along the same straight 30\,cm path, which ends as the target leaves the camera's view.
A rollout succeeds if, by then, the tote bag or case comes to rest in the bowl or on the plate and stays there without tipping over.
For \emph{grasping}, the object starts moving once the robot has executed its first action chunk.
For \emph{release}, the receptacle starts moving once the robot has lifted the tote bag off the table.
A motor pulls the moving object through a transparent wire at a constant 0.02 or 0.05\,m/s, and only one object moves at a time.
Each rollout starts with the arm and the objects reset to the same positions.

We first check the setup with static targets, then move the targets.
Under synchronous execution, we place the target of every task at the start, middle, and end of its path and test every latency.
Every static setting succeeds in all five rollouts, which confirms that the path, the object placement, and the tasks themselves work before the targets move.
We then move the targets at 0.02 and 0.05\,m/s under synchronous execution on every task, and under lazy and continuous asynchronous execution on \emph{release small}, the harder release task, to measure how much asynchronous execution recovers.
We set the latencies from MolmoAct2's measured median inference latency of 209\,ms: holding each action chunk until $L$ control steps after its observation, we test 267, 367, and 467\,ms ($L=8$, 11, and 14), all above the median so that nearly every chunk takes effect at the set latency, with continuous execution at 267\,ms ($H=L=8$).
Each reported result is based on five rollouts.

\textbf{Finding~1 holds.} Inference latency substantially reduces task success on moving targets, to an extent that depends on the task and target speed.
Unlike static targets, which never fail, moving targets lose success: at 0.02\,m/s, increasing latency from 267 to 467\,ms reduces \emph{grasping small} success from three to zero of five rollouts and \emph{release small} success only from five to three (Figure~\ref{fig:real-robot-sync}(a,c)).

\textbf{Finding~2 holds.} A task's latency sensitivity depends on its requirements for action precision and continuous motion.
At 0.02\,m/s and 467\,ms, \emph{grasping big} and \emph{release big} succeed in one and four of five rollouts, whereas \emph{grasping small} and \emph{release small}, which require more precision, succeed in zero and three (Figure~\ref{fig:real-robot-sync}).
A faster target requires faster sustained motion: at 267\,ms, \emph{release small} succeeds in all five rollouts at 0.02\,m/s but in two at 0.05\,m/s (Figure~\ref{fig:real-robot-sync}c).

\textbf{Finding~3 holds.} Asynchronous execution recovers task success, and continuous asynchronous execution recovers the most.
On \emph{release small} at 0.05\,m/s and 267\,ms, continuous execution succeeds in all five rollouts, lazy execution in four, and synchronous execution in two (Figure~\ref{fig:real-robot-modes}a).

\FloatBarrier
\section{Failure stages}
\label{app:failures}

Latency costs success at the stages where the robot interacts with the moving target, while the stages before and away from it hold (Table~\ref{tab:failure-stages}).
Each task is split into the stages an episode passes on the way to success, and each rate counts the episodes passing a stage among those that reached it.
In \emph{release}, the robot picks up the static object and lets go of it over the moving receptacle; pickup means both fingers hold the object and it rises at least 2\,cm, and release means both fingers then lose contact with it.
In \emph{set-down}, placement means the object rests on the moving surface with stable contact for 0.2\,s.
In \emph{grasping}, the robot picks up the moving object and then places it on a static surface.
In \emph{articulated-object manipulation}, the hand first touches the handle of the moving drawer, and opening means the drawer moves at least 1\,cm from its starting position.
On \emph{release}, 200\,ms synchronous latency lowers success after release from 80\% to 58\%, while pickup and release stay at or above 92\%.
On \emph{grasping}, the loss is at pickup of the moving object (85\% to 73\%); once the object is held, placing it on the static surface succeeds about as often as without latency.
Asynchronous execution recovers these stages on \emph{articulated-object manipulation} and partly on \emph{release}, but little on \emph{set-down} and \emph{grasping}.

\begin{table}[t]
  \centering
  \small
  \caption{Stage pass rates of $\pi_{0.5}$ (\%), each among the episodes reaching that stage, over all nine target speeds; $\triangleright$ marks stages at the moving target.}
  \label{tab:failure-stages}
  \begin{tabular}{@{}lcccc@{}}
  \toprule
  & \multicolumn{2}{c}{Synchronous} & Lazy & Continuous \\
  \cmidrule(lr){2-3}\cmidrule(lr){4-4}\cmidrule(l){5-5}
  Task / stage & 0\,ms & 200\,ms & 200\,ms & 200\,ms \\
  \midrule
  \multicolumn{5}{@{}l}{\emph{Release}} \\
  \quad Pickup & 100 & 100 & 100 & 100 \\
  \quad Release & 92 & 96 & 98 & 96 \\
  \quad $\triangleright$ Success after release & 80 & 58 & 69 & 72 \\
  \multicolumn{5}{@{}l}{\emph{Set-down}} \\
  \quad $\triangleright$ Placement & 90 & 65 & 67 & 70 \\
  \quad $\triangleright$ Success after placement & 72 & 66 & 65 & 67 \\
  \multicolumn{5}{@{}l}{\emph{Grasping}} \\
  \quad $\triangleright$ Pickup & 85 & 73 & 73 & 76 \\
  \quad Success after pickup & 94 & 91 & 89 & 90 \\
  \multicolumn{5}{@{}l}{\emph{Articulated-object manipulation}} \\
  \quad Handle contact & 97 & 98 & 91 & 92 \\
  \quad $\triangleright$ Opening $\geq$ 1\,cm & 66 & 48 & 63 & 62 \\
  \quad $\triangleright$ Success after opening & 47 & 38 & 52 & 54 \\
  \bottomrule
  \end{tabular}
\end{table}

\section{Additional related work}
\label{app:related}

Several model and system papers handle inference latency or execution speed within their own methods.
On the systems side, SmolVLA introduces an asynchronous inference stack that overlaps chunk prediction with execution and compares it with synchronous execution on three real-world tasks~\citep{shukor2025smolvla}, and VLASH conditions asynchronous inference on robot states rolled forward with the previous action chunk~\citep{tang2025vlash}.
On the policy-pacing side, SAIL executes imitation policies faster than their demonstrations and schedules actions around system latency~\citep{arachchige2025sail}, DemoSpeedup accelerates demonstrations by entropy-guided downsampling without modeling inference delay~\citep{guo2025demospeedup}, and TempoVLA conditions $\pi_{0.5}$ on a commanded speed~\citep{jing2026tempovla}.
\citet{akinola2021dynamicgrasping} predict object motion for dynamic grasping because computation delays the robot's response.

\end{document}